\documentclass[lettersize,journal]{IEEEtran}
\usepackage{amsmath,amsfonts}
\usepackage{algorithmic}
\usepackage{algorithm}
\usepackage{array}
\usepackage[caption=false,font=normalsize,labelfont=sf,textfont=sf]{subfig}
\usepackage{textcomp}
\usepackage{stfloats}
\usepackage{url}
\usepackage{verbatim}
\usepackage{graphicx}
\usepackage{cite}
\usepackage{multirow}
\usepackage{xcolor}
\usepackage{amssymb}
\usepackage{tabularray}
\usepackage{newfloat}
\usepackage{listings}
\usepackage{booktabs}
\renewcommand{\arraystretch}{1.3}

\begin{document}

\title{SPDA-SAM: A Self-prompted Depth-Aware Segment Anything Model for Instance Segmentation}

\author{
Yihan Shang,
Wei Wang,
Chao Huang, \IEEEmembership{Member, IEEE},
Junyu Dong,~\IEEEmembership{Member, IEEE} and
Xinghui Dong, \IEEEmembership{Member, IEEE}
\thanks{This study was supported in part by the National Natural Science Foundation
of China (NSFC) (No. 42576200) and in part by the Key Research
and Development Program of Shandong Province, China (No. 2024ZLGX06) (Corresponding author: Xinghui Dong).}
\thanks{Y. Shang, J. Dong and X. Dong are with the State Key Laboratory of Physical Oceanography and the Faculty of Information Science and Engineering, Ocean University of China, Qingdao, 266100. (e-mail: shangyihan@stu.ouc.edu.cn, dongjunyu@ouc.edu.cn, xinghui.dong@ouc.edu.cn).}
\thanks{W. Wang and C. Huang are with the School of Cyber Science and Technology, Shenzhen Campus of Sun Yat-sen University, Shenzhen, 518107. (e-mail: wangwei29@mail.sysu.edu.cn; huangch253@mail.sysu.edu.cn).
}

}



\maketitle

\begin{abstract}
Recently, Segment Anything Model (SAM) has demonstrated strong generalizability in various instance segmentation tasks. However, its performance is severely dependent on the quality of manual prompts. In addition, the RGB images that instance segmentation methods normally use inherently lack depth information. As a result, the ability of these methods to perceive spatial structures and delineate object boundaries is hindered. To address these challenges, we propose a Self-prompted Depth-Aware SAM (SPDA-SAM)\footnote{The source code and models will be made publicly available on the acceptance of the paper.} for instance segmentation. Specifically, we design a Semantic-Spatial Self-prompt Module (SSSPM) which extracts the semantic and spatial prompts from the image encoder and the mask decoder of SAM, respectively. Furthermore, we introduce a Coarse-to-Fine RGB-D Fusion Module (C2FFM), in which the features extracted from a monocular RGB image and the depth map estimated from it are fused. In particular, the structural information in the depth map is used to provide coarse-grained guidance to feature fusion, while local variations in depth are encoded in order to fuse fine-grained feature representations. To our knowledge, SAM has not been explored in such self-prompted and depth-aware manners. Experimental results demonstrate that our SPDA-SAM outperforms its state-of-the-art counterparts across eleven different data sets. These promising results should be due to the guidance of the self-prompts and the compensation for the spatial information loss by the coarse-to-fine RGB-D fusion operation. 
\end{abstract}


\section{Introduction}
As a fundamental computer vision task, Instance Segmentation aims at accurately identifying and delineating each individual object within an image. It has been widely applied in various domains, such as autonomous driving \cite{zhou2020joint}, medical image analysis \cite{xu2025lightsam}, and remote sensing detection \cite{su2020hq,dong2025two}. However, the existing instance segmentation models, which were trained using a relatively small data set, usually encounter challenges when applied to a different data set because of the domain-shift issue \cite{hsu2021darcnn}.


Recently, Segment Anything Model (SAM) \cite{Kirillov_2023_ICCV} has demonstrated strong generalizability across different domains, due to the large-scale data set, i.e., SA-1B \cite{Kirillov_2023_ICCV} on which it was pre-trained. As a representative promptable model, SAM has shown impressive performance in various instance segmentation tasks and has been increasingly adopted in diverse downstream applications \cite{chen2024rsprompter,wang2023detect,hou2025hypsam}. However, the performance of it is affected by the quality of prompts \cite{tang2025ssp}. This dependency not only increases the cost of human intervention, but also limits the deployment and automaticity of the model \cite{chen2023sam} in practical scenarios.

\begin{figure}
  \centering
  \includegraphics[width=1.0\linewidth]{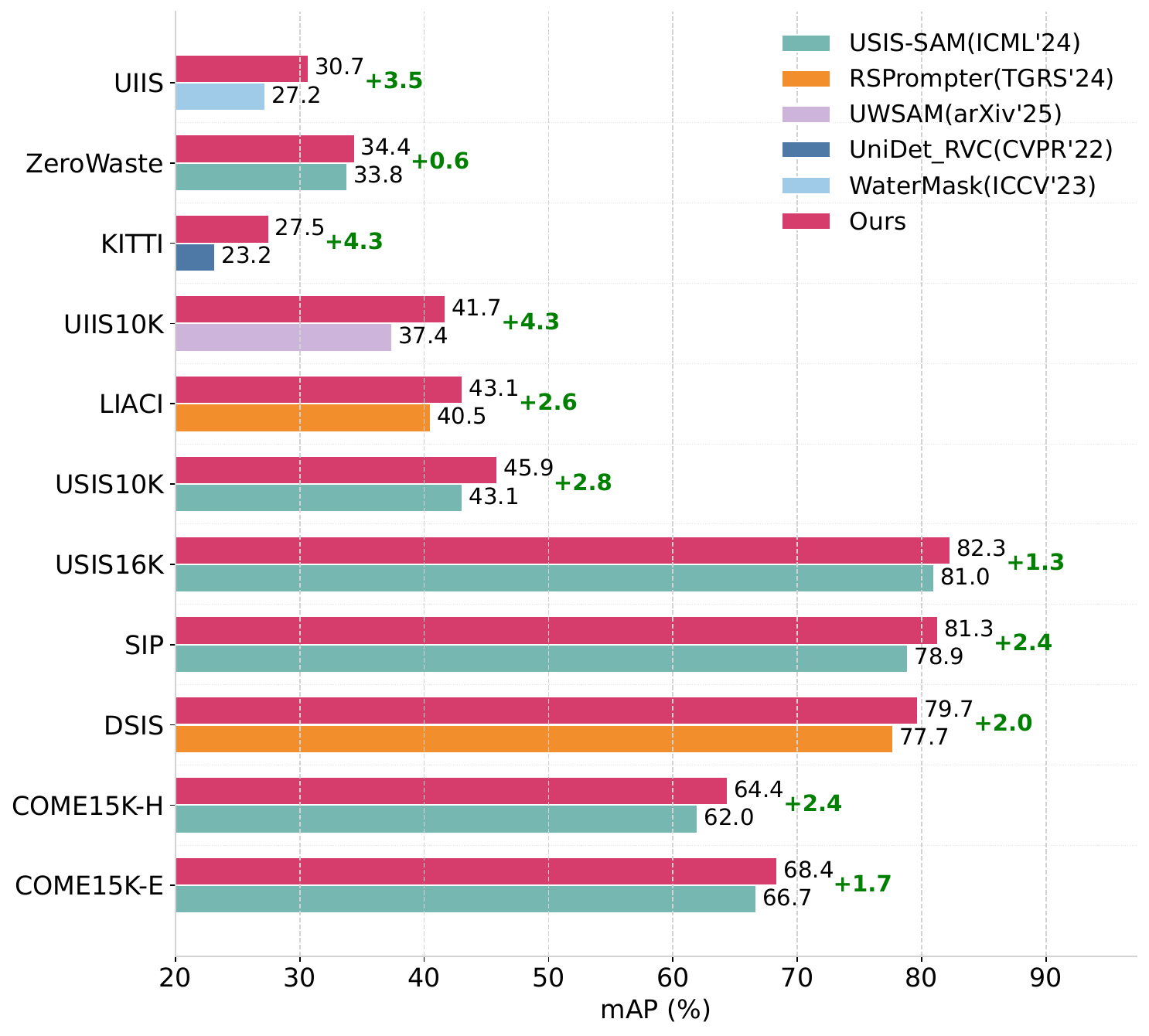}
  \caption{Comparison between five state-of-the-art methods and our SPDA-SAM across 11 instance segmentation data sets.}
  \label{fig:head}
\end{figure}

In addition, depth information is useful for instance segmentation because it encoders rich geometric cues which represent object shapes, spatial positions, and relative sizes in the three-dimensional space. However, the monocular RGB images that existing instance segmentation methods normally utilize inevitably lose depth information during the imaging procedure, which impairs the ability of these methods to understand spatial structures and object boundaries. 

\IEEEpubidadjcol

To address the aforementioned challenges, we propose a Self-prompted Depth-Aware SAM, referred to as SPDA-SAM, for instance segmentation. This network eliminates the manual prompts provided by humans and incorporates depth information into image representations. For the purpose of adapting to a specific domain, the SPDA-SAM is fine-tuned using LoRA \cite{hu2022lora} with a relatively small data set.

To be specific, we first design a Semantic-Spatial Self-prompt Module (SSSPM), which extracts semantic and spatial prompts from the image encoder and mask decoder of SAM, respectively. The former captures semantic representations to utilize high-level contextual information, while the latter leverages the sensitivity of the mask decoder to object boundaries to enhance the detail modeling ability. As a result, the adaptive segmentation guidance and autonomous perception capabilities of the model can be obtained. 

We then introduce a Coarse-to-Fine RGB-D Fusion Module (C2FFM) in which the features extracted from the depth map that we estimate using a pre-trained monocular depth estimation model are fused with those extracted from an RGB image. Contours extracted from the depth map are used to guide the attention distribution at the coarse level. On the other hand, local depth gradients are used to encode boundary details at the fine level, which enable the more precise instance segmentation results to be produced. The entire network can be end-to-end trained without human intervention. 

To our knowledge, SAM has not been explored in such self-prompted and depth-aware manners. Our main contributions can be summarized as threefold. 
\begin{itemize}

\item We introduce a Self-prompted Depth-Aware SAM, i.e., SPDA-SAM, for the instance segmentation task. Due to the powerful generalizability of SAM, our network can adapt to a novel scenario by fine-tuning it using a small number of images. The SPDA-SAM gets rid of the user-input prompts and injects depth information into image representations. As shown in Fig. \ref{fig:head}, our method always outperforms the state-of-the-art baseline on each of the eleven data sets.

\item We design a Semantic-Spatial Self-prompt Module (SSSPM), which exploits both the semantic and spatial prompts generated automatically. Therefore, human intervention is not required.

\item We propose a Coarse-to-Fine RGB-D Fusion Module (C2FFM) which progressively fuses the features extracted from the depth map that we estimate from a single RGB image and those extracted from this image. As a result, the discriminatory power of the encoder becomes stronger.

\item We conduct comprehensive evaluations on eleven publicly available data sets. The results provide the community with a set of benchmarks.

\end{itemize}

The remainder of this paper is organized as follows. In Section II, we review the related work. The proposed SPDA-SAM method is described in Section III. In Section IV, our experimental setup and results and the ablation study are reported. Finally, we draw our conclusion in Section V.


\section{Related Work}

\subsection{Instance Segmentation}
Instance segmentation requires the model not only to distinguish between different object categories, but also to differentiate individual instances within the same category. 
Early approaches were usually adopted on top of a two-stage framework, such as Mask R-CNN \cite{he2017mask}, Cascade Mask R-CNN \cite{cai2019cascade} and HTC \cite{chen2019hybrid}. These approaches typically followed a two-step paradigm, which first generates region proposals and then performs mask prediction. Although those approaches achieved a high accuracy, they often underwent heavy computational overhead, which limited their applicability in real-time or resource-constrained scenarios. Moreover, the reliance on hand-crafted region proposal generators further impaired the adaptability of the model to novel domains. 

To improve efficiency and performance, one-stage instance segmentation methods were subsequently developed, for example, YOLACT \cite{bolya2019yolact}, BlendMask \cite{chen2020blendmask}, CondInst \cite{tian2020conditional}, EmbedMask\cite{ying2019embedmask} and SOLOv2 \cite{wang2020solov2}. These methods directly generated masks without relying on region proposals, which significantly accelerated the inference process. However, they often encountered challenges in handling overlapping instances and complex object shapes, which might lead to low segmentation accuracy in densely populated or cluttered scenes. Recently, Transformer-based approaches, e.g., Mask2Former \cite{cheng2022masked}, were developed on top of attention mechanisms, because they leveraged global context information to better handle complex scenes and improve the discrimination between adjacent instances.

However, existing methods often depended heavily on the distributions of training data and tended to suffer performance degradation when applied to complex or domain-specific images, e.g., underwater scenarios, medical imagery and remote sensing images. This issue hinders the deployment of them in practical applications and has motivated a growing interest in foundational vision models. Therefore, we were motivated to perform instance segmentation tasks on top of such a model.

\subsection{Segment Anything Models}


As a foundation model in computer vision, Segment Anything Model (SAM) \cite{Kirillov_2023_ICCV} has demonstrated remarkable generalizability in class-agnostic segmentation tasks on natural images. This model generates segmentation masks based on a prompt-driven mechanism. It has been extended to a variety of specialized segmentation scenarios, including medical imaging \cite{chen2024ma}, remote sensing \cite{chen2024rsprompter} and underwater imagery \cite{lian2024diving}. However, its performance often degraded greatly in such domain-specific tasks, due to the substantial distribution gap between the pre-training data and target domains. To make the model better adapt to the target domain, we thus used a small data set in this domain to fine-tune the pre-trained SAM.

In addition, the severe reliance of SAM on explicit external prompts for target localization poses at least two challenges \cite{Kirillov_2023_ICCV,lian2024diving}. First, high-quality prompt acquisition typically requires additional annotation efforts, which can be prohibitive when dealing with large-scale data sets. Second, the model exhibits limited robustness to input perturbations. Minor variations in the location or the form of the prompts can significantly affect the segmentation results \cite{Kirillov_2023_ICCV,chen2024rsprompter}. In this case, SAM also lacks the ability to automatically segment instances within domain-specific applications, even though it has demonstrated excellent performance on natural images. To address this issue, we proposed a self-prompted module, which eliminates human intervention during prompt generation.


\subsection{Prompt Learning} 

In recent years, a new learning paradigm, i.e., pre-training and prompting, has gradually become mainstream with the emergence of foundation models \cite{jia2022visual}. This paradigm eliminates the need for fine-tuning by reformulating downstream tasks through the design of input prompts, thereby aligning them more closely with the pre-training objectives. As a representative technique, Contrastive Language-Image Pre-training (CLIP) \cite{radford2021learning} leveraged textual prompts for cross-modal understanding. On the other hand, Kirillov et al. \cite{Kirillov_2023_ICCV} made the first effort on developing a foundation model for image segmentation, i.e., SAM, on top of the prompt-based mechanism. However, the prompts of SAM require human intervention. To tackle this problem, we introduced a Semantic-Spatial Self-prompt Module (SSSPM), which utilize both the semantic and spatial
prompts generated automatically.

\begin{figure*}
  \centering
  \includegraphics[width=1.0\linewidth]{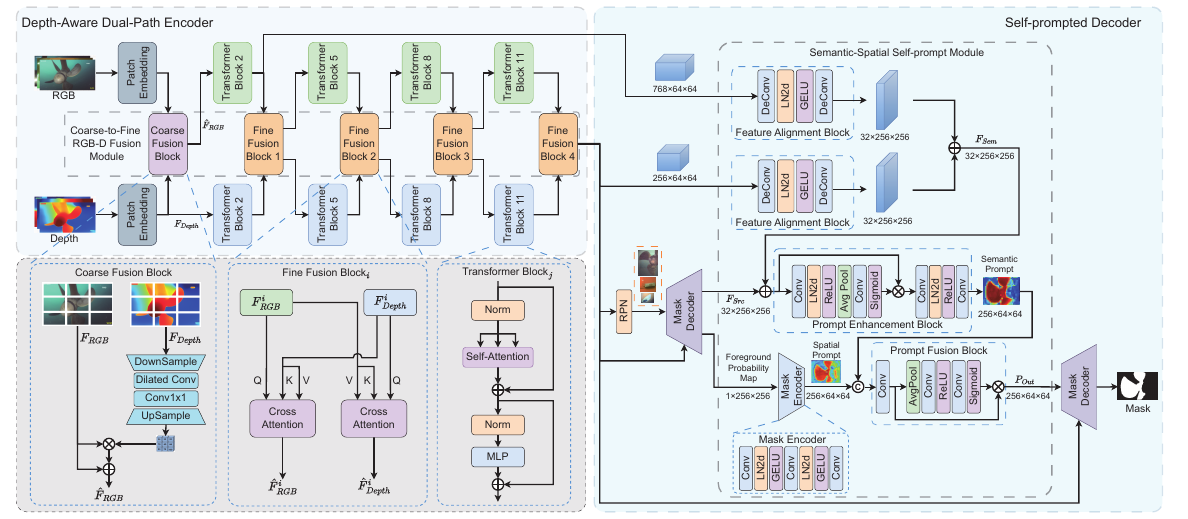}
  
  \caption{
  The architecture of the Self-prompted Depth-Aware SAM (SPDA-SAM), which consists of a depth-aware dual-path encoder and a self-prompted decoder. The encoder contains an RGB encoding path, a depth encoding path and a Coarse-to-Fine RGB-D Fusion Module (C2FFM). The decoder includes a mask decoder and a Semantic-Spatial Self-prompt Module (SSSPM). 
  }
  \label{fig:architecture}
\end{figure*}

\subsection{RGB-D Based Image Segmentation}

In contrast to the RGB images that image segmentation methods normally utilize, the joint use of RGB images and depth maps (i.e., RGB-D) allows for exploitation of the spatial characteristics encoded in the depth maps. 
Previous studies have made progress by incorporating depth characteristics into feature representations \cite{ying2022uctnet}. However, effective feature extraction from depth maps and fusion with RGB image features remain challenges. This dilemma should be attributed to the significant difference between depth maps and RGB images in terms of data distributions, numerical range, and semantic representation. 

Depth-aware convolutions, such as depth-aware CNNs \cite{yang2023pixel}, have been introduced. They incorporated specially designed convolutional kernels or attention mechanisms to enable the network to perceive and exploit depth information. On the other hand, dual-branch encoders \cite{zhang2023feature} were proposed, in which RGB images and depth maps were processed by two individual encoders, to extract modality-specific features. The two sets of features were then fused \cite{ying2022uctnet}. However, these methods usually relied on the simple concatenation or shallow fusion operations. As a result, they failed to fully exploit the structural semantics embedded in the depth maps. In contrast, we designed a C2FFM, which is able to progressively fuse the features extracted
from an image and the associated depth map, to improve the discriminatory power of the encoder.

\section{Methodology}

To alleviate the reliance of SAM \cite{Kirillov_2023_ICCV} on manual prompts and address the absence of depth information in instance segmentation, we propose a Self-prompted Depth-Aware SAM, namely, SPDA-SAM. Due to the excellent generalizability of SAM, this network can be used for different instance segmentation tasks. The architecture of the SPDA-SAM is shown in Fig. \ref{fig:architecture}. As can be seen, our SPDA-SAM contains a Depth-Aware Dual-Path Encoder and a Self-prompted Decoder. The former includes an RGB Encoding Path, a Depth Encoding Path and a Coarse-to-Fine RGB-D Fusion Module (C2FFM), while the latter comprises a Mask Decoder and a Semantic-Spatial Self-prompt Module (SSSPM).

\subsection{Depth-Aware Dual-Path Encoder}

Within the encoder, the RGB encoding path and the depth encoding path are built on top of the backbone of SAM \cite{Kirillov_2023_ICCV}. 
The C2FFM include a Coarse Fusion Block and four Fine Fusion Blocks. The former is applied to the output of the Patch Embedding Block, while the latter are applied to the output of the 2nd, 5th, 8th and 11th Transformer Blocks individually.

\subsubsection{RGB Encoding Path}

The RGB Encoding Path is initialized using the weights of the backbone of SAM. This path is used to extract features from RGB images. To adapt the RGB encoding path to a new scenario, we use LoRA \cite{hu2022lora} to fine-tune it.

\subsubsection{Depth Encoding Path}

This path is also initialized using the weights of the backbone of SAM, which aims to extract features from the depth map. LoRA \cite{hu2022lora} is also used to fine-tune the depth encoding path.

\subsubsection{C2FFM}

The C2FFM progressively performs the coarse-to-fine fusion operation on the features extracted from both the RGB image and the depth map across different stages, enabling effective utilization of the complementary information encoded in the two sets of data. A Coarse Fusion Block and four Fine Fusion Blocks are comprised of the C2FFM.

\textbf{\textit{Coarse Fusion Block}} 
The Coarse Fusion Block aims to integrate depth-guided structural features into RGB features at a coarse resolution, enabling the fusion of complementary modalities to enhance global context and boundary-aware representation. 
To achieve this goal, we first apply a downsampling operation, a dilated convolution and a Relu activation function to the patch embeddings of the depth map $F_{Depth}$, enlarging the receptive field and extracting multi-scale contextual information. Subsequently, a 1$\times$1 convolution is used to reduce the dimensionality of channels and fuse the features, followed by a upsampling operation to restore the resolution to match the feature maps $F_{RGB}$ extracted from the RGB image. 

The resultant feature maps are passed through a Sigmoid activation function and are then used to modulate $F_{RGB}$ through an element-wise multiplication operation. This operation highlights the regions with strong depth cues. A residual connection further fuses the original RGB features, yielding the fused output $\hat{F}_{RGB}$. The computation of the Coarse Fusion Block can be expressed as:
\begin{equation}
F_{Dilated}=DilConv(Down(F_{Depth})),
\end{equation}
\begin{equation}
Attn=Sigmoid(Up(Conv_{1\times 1}(F_{Dilated}))), 
\end{equation}
\begin{equation}
\hat{F}_{RGB}=F_{RGB}+Attn\odot F_{RGB},
\end{equation}
where $\sigma(\cdot)$ is the Sigmoid activation function, $\varphi (\cdot)$ denotes a sequence of downsampling, dilated convolution, and ReLU activation applied to the depth feature maps, and $\odot$ represents the element-wise multiplication operation.

\textbf{\textit{Fine Fusion Block}} This block is used to exploit the complementary actions of multi-modal features in local details. 
In total, four Fine Fusion Blocks are applied to the output of the 2nd, 5th, 8th and 11th Transformer Blocks, respectively. This hierarchical design enables the network to align multi-modal information across different levels of semantic abstraction. As a result, the network captures the complementary relationship between local fine-grained details and global contextual semantics. Regarding a Fine Fusion Block ${i}$, we first flatten the input feature maps $F_{RGB}^{i}$ and $F_{Depth}^{i}$ into sequential representations:
\begin{gather}
    S_{RGB}^{i}=Flatten(F_{RGB}^{i})\in \mathbb{R}^{B\times N\times C}, \\
    S_{Depth}^{i}=Flatten(F_{Depth}^{i})\in \mathbb{R}^{B\times N\times C},
\end{gather}
where $N=H\times W$ represents the length of the sequences. 

Two cross-attention blocks are applied, in which $S_{RGB}^{i}$ and $S_{Depth}^{i}$ are used as the query, respectively. Taking the case where $S_{RGB}^{i}$ is used as the query and $S_{Depth}^{i}$ is utilized as the key and value, we can obtain $\hat{F}_{RGB}^{i}$. This process can be formulated as:
\begin{gather}
Q,K,V=W_{Q}S_{RGB}^{i},W_{K}S_{Depth}^{i},W_{V}S_{Depth}^{i}, \\
\hat{F}_{RGB}^{i}=Reshape(Softmax(\frac{QK^{T}}{\sqrt{d_{k}} }V )),
\end{gather}
where $Reshape(\cdot)$ denotes the operation which transforms the sequence back to its original shape. Similarly, we obtain a set of feature maps $\hat{F}_{Depth}^{i}$.

\subsection{Self-prompted Decoder}

Compared with the mask decoder of the original SAM \cite{Kirillov_2023_ICCV} which requires external prompts, we propose a Self-prompted Decoder. This decoder contains a mask decoder and an SSSPM. The self-prompted decoder
leverages both the internal visual features extracted using the depth-aware dual-path encoder and the mask that the mask decoder produces to generate self-guided prompts, which are able to direct the segmentation process. The proposed self-prompt mechanism enables accurate object localization and segmentation without human intervention.

\subsubsection{Mask Decoder}

The mask decoder, which has the same structure as that of SAM but is randomly initialized, is used twice. At the first time, it receives the initial prompts generated by the Region Proposal Network (RPN) \cite{ren2016faster} and the output of the last fine fusion block. The result is a coarse segmentation mask. This process roughly localizes the target region. At the second time, the mask decoder takes the semantically and spatially enriched prompts that the proposed SSSPM produces and the output of the last fine fusion block. As a result, the mask boundaries and structural details are further refined.

\subsubsection{SSSPM}

The SSSPM aims to automatically generate high-quality prompts without human intervention. The prompts comprises semantic and spatial prompts, which are visualized in Fig. \ref{fig:architecture} by computing the dot-product similarity between the center vector and surrounding vectors.
The SSSPM consists of four main components, including a Feature Alignment Block, a Mask Encoder, a Prompt Enhancement Block and a Prompt Fusion Block.

\textbf{\textit{Feature Alignment Block}} This block facilitates the integration of multi-level semantic cues. We select two representative layers from the depth-aware dual-path encoder, including a lower-level block (i.e., Transformer Block 2) which captures local details and a higher-level block (i.e., Fine Fusion Block 4) which encodes high-level abstract semantics. The two sets of features produced by both layers differ significantly in resolution and channel dimensions. To achieve an effective fusion operation, each set of features is passed through a Feature Alignment Block, which consists of a deconvolution \cite{long2015fully}, a layer normalization layer, a GELU activation function and a second deconvolution. This block spatially upsamples and channel-aligns the two sets of features and transforms them into a unified resolution and embedding space. The two sets of resultant features are fused via an element-wise addition operation. The result is a set of semantic features $F_{Sem}$, which contain rich semantic information and provide strong guidance for the subsequent prompt enhancement block.

\textbf{\textit{Prompt Enhancement Block}} The Prompt Enhancement Block aims to enrich the features $F_{Src}$ that the mask decoder produces, which have already been integrated with the initial prompt information. We fuse $F_{Src}$ with the semantic features $F_{Sem}$ via an element-wise addition operation. The fused features are then passed through the prompt enhancement block, which contains two convolutional layers, together with a layer normalization operation, a ReLU activation function and a global average pooling operation. A Sigmoid activation is used to obtain an attention map, which modulates the fused features. Finally, two convolutional layers are used to reduce the spatial size and increase the channel dimension. As a result, the semantic prompt is derived.

\textbf{\textit{Mask Encoder}} The Mask Encoder receives a coarse foreground probability map that the mask decoder generates. The map is transformed into dense embeddings, which have been spatially aligned with the image embeddings. The mask encoder contains two strided \(2 \times 2\) convolutional layers. Each layer is followed by a layer normalization operation and a GELU activation function. A \(1 \times 1\) convolution is finally used to perform the dimensional projection operation. The resultant dense embeddings encode spatial priors and serve as the spatial prompt \(P_{Spatial}\).

\textbf{\textit{Prompt Fusion Block}} This block is designed to exploit the complementary relationship between the spatial prompt $P_{Spatial}$ and the semantic prompt $P_{Semantic}$. Specifically, they are first concatenated along the channel dimension, followed by a convolutional layer, to obtain a fused prompt $P_{Fused}$. For the sake of extracting global contextual information from $P_{Fused}$, we apply a global average pooling operation to it. This process summarizes the response of each channel, which facilitates the generation of an attention map in order to selectively emphasize informative channels. The pooled features are passed through two convolutional layers with a ReLU activation function in between, followed by a Sigmoid activation function. The result is an attention map $Attn$, which is then used to modulate $P_{Fused}$ via an element-wise multiplication operation. 

The final prompt that the SSSPM generates, $P_{Out}$, can be formulated as:
\begin{gather}
P_{Fused}=Conv(Concat(P_{Semantic},P_{Spatial})), \\
Attn=\sigma (Conv(ReLU(Conv(GAP(P_{Fused}))))), \\
P_{Out}=P_{Fused}\odot Attn,
\end{gather}
where $Concat(\cdot)$ denotes the channel-wise concatenation operation, $GAP(\cdot)$ stands for the global average pooling operation, and $\sigma(\cdot)$ represents the Sigmoid activation function. $P_{Out}$ is fed into the mask decoder, together with the output of the last fine fusion block. This process allows the mask decoder to more accurately distinguish between target regions with ambiguous boundaries or high semantic similarity. Consequently, the accuracy of instance segmentation can be improved.

\subsection{Loss Function}  
In addition to the loss functions of Region Proposal Network (RPN) \cite{ren2016faster}, which include the object classification and bounding box regression loss functions, we apply two pixel-wise cross-entropy loss functions to supervise both the coarse and refined masks generated by applying the mask decoder twice. Finally, the combined loss function $\mathcal{L}$ can be written as
\begin{gather}
\mathcal{L} = \lambda_{1} \cdot  \mathcal{L}_{RPN} + \lambda_{2} \cdot \mathcal{L}_{Coarse} + \lambda_{3} \cdot \mathcal{L}_{Refined},
\end{gather}
where $\mathcal{L}_{RPN}$ denotes the RPN loss, and $\mathcal{L}_{Coarse}$ and $\mathcal{L}_{Refined}$ stands for the cross-entropy losses used for the coarse and refined masks, respectively. In our experiments, all the three weights were set to 1 (i.e., $\lambda_{1} = \lambda_{2} = \lambda_{3} = 1$). This multi-stage supervision strategy ensures the consistency between coarse localization and final segmentation, which improves the quality of the final mask.

\section{Experiments}

In this section, we first describe the setup used in our experiments. Then we report the results obtained in the main experiments and ablation study.

\subsection{Experimental Setup}

Here, we introduce the data sets and evaluation metrics that we used and the implementation details of the experiments.

\subsubsection{Data Sets}

We evaluated the proposed method on eleven publicly available data sets. Specifically, six terrestrial data sets, including COME15K-E \cite{zhang2021rgb}, COME15K-H \cite{zhang2021rgb}, DSIS \cite{pei2024calibnet}, SIP \cite{fan2020rethinking}, KITTI \cite{geiger2013vision} and ZeroWaste \cite{bashkirova2022zerowaste}, and five UIIS data sets, including LIACI \cite{waszak2022semantic}, UIIS \cite{lian2023watermask}, USIS10K \cite{lian2024diving}, UIIS10K \cite{li2025uwsam} and USIS16K \cite{hong2025usis16k}, were used. We always followed the original splitting for each data set, in which the data set was divided into training, validation and test sets. Regarding the data sets with instance-level annotations, we directly used them as the ground-truth data. In terms of the data sets which only contained semantic annotations, we used connected component labeling to convert semantic segmentation masks to instance-level annotations.

The pre-processing strategy proposed in Depth Anything v2 \cite{yang2024depth} was used to perform monocular depth estimation on RGB images. This process aimed to recover the depth information lost during the imaging procedure.


\subsubsection{Evaluation Metrics}

To evaluate the performance of the proposed method on instance segmentation tasks, we utilized the widely used mean Average Precision ($mAP$) metric \cite{lin2014microsoft}, as well as $AP_{50}$ and $AP_{75}$, following existing studies.

\subsubsection{Implementation Details}

We adopted the ViT-B \cite{dosovitskiy2020image} model in the pre-trained SAM \cite{Kirillov_2023_ICCV} to initialize the RGB and depth encoding paths individually, and applied LoRA\cite{hu2022lora} to the QKV projection layers with a rank of 16, a scaling factor of 32, a dropout rate of 0.05, and no bias term. The proposed method was implemented using PyTorch and the MMDetection framework \cite{chen2019mmdetection}. During the training process, we used the AdamW optimizer with an initial learning rate of 0.0002 and a weight decay of 0.05. To enable efficient and stable optimization, we designed a two-stage learning rate scheduling strategy. Specifically, we employed a linear warm-up scheme for the first 50 iterations, which gradually increased the learning rate to the target value. Subsequently, we applied a cosine annealing learning rate schedule, to progressively decay the learning rate throughout the training process, which eventually reduced this rate to 0.001 times the base rate. Additionally, we used the Automatic Mixed Precision (AMP) training to accelerate convergence and reduce memory consumption. For all data sets, the batch size was set to 2 during the training process. All experiments were conducted on either two NVIDIA 3080Ti GPUs or a single NVIDIA 4090 GPU.

\subsection{Main Experiments}

We tested our method along with baselines on the eleven data sets. We will report the results as follows.

\subsubsection{COME15K-E, COME15K-H, DSIS and SIP}

We first evaluated the SPDA-SAM, together with 22 baseline approaches, on four salient instance segmentation data sets, including COME15K-E \cite{zhang2021rgb}, COME15K-H \cite{zhang2021rgb}, DSIS \cite{pei2024calibnet} and SIP \cite{fan2020rethinking}. As reported in Table \ref{tab:salient_segmentation}, the SPDA-SAM always achieved the superior performance to its counterparts. On the COME15K-E data set, our method outperformed the second-best approach by 1.7, 2.9 and 3.6 in terms of the $mAP$, $AP_{50}$ and $AP_{75}$ metrics, respectively. Regarding the COME15K-H data set, the three values were 2.4, 3.4 and 3.8, respectively. Using the DSIS data set, our method achieved the improvements of 2.0, 1.3 and 0.8 in terms of the three metrics, respectively, compared to the second-best approach. With regard to the SIP data set, the three values were 2.4, 1.5 and 1.8, respectively. These results demonstrate the robustness and generalizability of our SPDA-SAM across different salient instance segmentation scenarios.

\begin{table*}
\centering
\setlength{\tabcolsep}{5pt}
\caption{Comparison between our method and 22 baselines on the COME15K-E \cite{zhang2021rgb}, COME15K-H \cite{zhang2021rgb}, DSIS \cite{pei2024calibnet} and SIP \cite{fan2020rethinking} data sets. The best and second-best results are highlighted in \textbf{bold} and \underline{underline} fonts, respectively. This continues with following tables.}
\label{tab:salient_segmentation}
\begin{tabular}{c|c|ccc|ccc|ccc|ccc} 
\toprule
\multirow{2}{*}{Method} & \multirow{2}{*}{Venue} & \multicolumn{3}{c|}{COME15K-E}                & \multicolumn{3}{c|}{COME15K-H}                & \multicolumn{3}{c|}{DSIS}                     & \multicolumn{3}{c}{SIP}                        \\
\cline{3-14}
                        &                         & $mAP$         & $AP_{50}$       & $AP_{75}$       & $mAP$         & $AP_{50}$       & $AP_{75}$       & $mAP$         & $AP_{50}$       & $AP_{75}$       & $mAP$         & $AP_{50}$       & $AP_{75}$        \\ 
\hline\hline
S4Net\cite{fan2019s4net}                   & CVPR19                  & 43.7          & 68.0          & 52.5          & 37.1          & 60.9          & 43.2          & 58.3          & 81.9          & 71.8          & 49.6          & 76.0          & 63.7           \\
RDPNet\cite{wu2021regularized}                  & TIP21                   & 49.8          & 72.2          & 59.5          & 42.1          & 65.2          & 49.7          & 66.1          & 87.2          & 80.1          & 59.0          & 80.1          & 74.1           \\
OQTR\cite{pei2022transformer}                    & TMM22                   & 48.7          & 71.1          & 58.3          & 42.7          & 65.9          & 50.5          & 63.1          & 85.9          & 77.0          & 59.9          & 83.1          & 76.3           \\
OSFormer\cite{pei2022osformer}                & ECCV22                  & 53.0          & 71.9          & 61.3          & 45.8          & 66.3          & 52.4          & 67.9          & 86.3          & 78.3          & 63.2          & 80.5          & 74.6           \\
Mask R-CNN \cite{he2017mask}              & ICCV17                  & 48.8          & 71.2          & 58.6          & 42.2          & 65.7          & 50.8          & 65.6          & 86.8          & 80.2          & 57.9          & 79.8          & 73.3           \\
Mask Scoring R-CNN\cite{pei2022transformer}                & CVPR19                  & 49.7          & 70.4          & 58.9          & 42.3          & 63.8          & 50.0          & 66.8          & 87.3          & 80.3          & 60.0          & 79.8          & 73.3           \\
Cascade R-CNN\cite{cai2019cascade}           & TPAMI19                 & 49.4          & 69.8          & 58.7          & 41.9          & 64.3          & 49.8          & 66.4          & 87.3          & 79.9          & 59.1          & 79.1          & 73.6           \\
CenterMask\cite{lee2020centermask}              & CVPR20                  & 49.6          & 71.6          & 59.0          & 42.5          & 65.2          & 51.0          & 65.7          & 87.6          & 79.7          & 57.6          & 79.8          & 72.3           \\
HTC\cite{chen2019hybrid}                     & CVPR19                  & 51.4          & 73.7          & 61.0          & 44.1          & 68.0          & 52.1          & 67.5          & 87.6          & 81.1          & 60.0          & 81.4          & 74.9           \\
YOLACT\cite{bolya2019yolact}                  & ICCV19                  & 48.1          & 70.7          & 56.2          & 41.4          & 66.0          & 48.5          & 62.4          & 84.1          & 73.3          & 62.0          & 82.3          & 74.7           \\
BlendMask\cite{chen2020blendmask}               & CVPR20                  & 48.1          & 70.6          & 56.9          & 41.0          & 64.8          & 48.5          & 65.5          & 86.9          & 78.0          & 55.5          & 77.3          & 69.7           \\
CondInst\cite{tian2020conditional}                & ECCV20                  & 49.6          & 72.0          & 59.5          & 42.8          & 66.4          & 51.0          & 65.1          & 86.8          & 79.1          & 58.6          & 79.4          & 73.3           \\
SOLOv2\cite{wang2020solov2}                  & TPAMI21                 & 51.1          & 71.6          & 59.9          & 45.1          & 66.3          & 52.8          & 67.4          & 87.0          & 80.4          & 63.4          & 80.7          & 74.8           \\
QueryInst\cite{fang2021instances}               & CVPR21                  & 51.5          & 73.1          & 61.1          & 43.9          & 67.5          & 51.8          & 67.9          & 87.5          & 80.6          & 61.3          & 81.0          & 75.1           \\
SOTR\cite{guo2021sotr}                    & ICCV21                  & 50.7          & 70.4          & 59.0          & 43.5          & 65.1          & 50.3          & 68.2          & 86.9          & 79.3          & 61.6          & 78.4          & 73.0           \\
Mask Transfiner\cite{ke2022mask}         & CVPR22                  & 48.7          & 71.0          & 56.3          & 40.7          & 64.5          & 47.2          & 67.5          & 87.6          & 80.4          & 57.8          & 78.9          & 70.3           \\
SparseInst\cite{cheng2022sparse}              & CVPR22                  & 51.3          & 71.9          & 58.9          & 43.1          & 65.1          & 48.5          & 65.0          & 85.0          & 75.7          & 62.8          & 81.3          & 75.3           \\
Mask2Former \cite{cheng2022masked}             & CVPR22                  & 51.5          & 67.3          & 57.6          & 44.1          & 62.4          & 49.4          & 68.3          & 85.1          & 76.7          & 66.6          & 79.3          & 75.1           \\
PolySnake\cite{feng2024recurrent}                     & TCSVT24                   & 48.2          & 66.8          & 58.9          & 42.6          & 62.1          & 50.4          & 64.9          & 84.2          & 74.9          & 57.4          & 77.6          & 72.0        \\
CalibNet\cite{pei2024calibnet}                & TIP24                   & 58.0          & 75.8          & 65.6          & 50.7          & 70.4          & 57.3          & 69.3          & 87.8          & 81.6          & 72.1          & 86.6          & 82.9           \\ 
\hline
RSPrompter\cite{chen2024rsprompter}              & TGRS24                  & 65.0          & 81.6          & \underline{70.8}          & 59.9          & 78.6          & 65.0          & \underline{77.7}          & \underline{92.5}          & \underline{86.6}          & 78.8          & \underline{90.1}          & \underline{87.1}           \\
USIS-SAM\cite{lian2024diving}                & ICML24                  & \underline{66.7}          & \underline{81.7}          & 70.6          & \underline{62.0}          & \underline{78.7}          & \underline{65.5}          & 68.6          & 84.3          & 74.8          & \underline{78.9}          & 89.3          & 85.0           \\
\textbf{SPDA-SAM (Ours)}        & -                       & \textbf{68.4} & \textbf{84.6} & \textbf{74.4} & \textbf{64.4} & \textbf{82.1} & \textbf{69.3} & \textbf{79.7} & \textbf{93.8} & \textbf{87.4} & \textbf{81.3} & \textbf{91.6} & \textbf{88.9}  \\
\bottomrule
\end{tabular}
\end{table*}

\subsubsection{KITTI}

We further evaluated the proposed SPDA-SAM on the KITTI \cite{geiger2013vision} data set to verify its generalizability on real-world driving scenes. Since the official KITTI evaluation server only reports the $mAP$ and $AP_{50}$ metrics, we present the values of these metrics for comparison.
The results obtained using our SPDA-SAM and five baselines on the KITTI data set are reported in Table \ref{tab:kitti}. As can be seen, our SPDA-SAM achieved the best performance among the six methods, reaching an $mAP$ value of 27.5 and an $AP_{50}$ value of 49.2. Compared to the second-best approach, i.e., UniDet RVC \cite{zhou2022simple}, our method outperformed it by 4.3 and 0.1 in terms of $mAP$ and $AP_{50}$, respectively.

\begin{table}
\centering
\caption{Comparison between our method and five baseline methods on the KITTI \cite{geiger2013vision} data set.}
\label{tab:kitti}
\begin{tabular}{c|c|cc} 
\toprule
Method                                              & Venue     & $mAP$         & $AP_{50}$      \\ 
\hline\hline
UDeer DIS++ \cite{dong2023pep}                                         & ResNet-50 & 16.4          & 28.8           \\
CenterPoly \cite{perreault2021centerpoly}                                          & ResNet-50 & 8.7           & 26.7           \\
UniDet RVC \cite{zhou2022simple}                                         & ResNet-50 & \underline{23.2}  & \underline{49.1}   \\ 
\hline
RSPrompter \cite{chen2024rsprompter}                                         & ViT-Base  & 17.2         & 38.1          \\
USIS-SAM \cite{lian2024diving}                                           & ViT-Huge  & 18.8          & 41.5           \\
\textbf{\textbf{\textbf{\textbf{SPDA-SAM (Ours)}}}} & ViT-Base  & \textbf{27.5} & \textbf{49.2}  \\
\bottomrule
\end{tabular}
\end{table}

\subsubsection{LIACI}

To comprehensively evaluate the performance of our SPDA-SAM on the LIACI \cite{waszak2022semantic} data set, we compared it with ten state-of-the-art instance segmentation approaches, including two-stage methods \cite{he2017mask} and one-stage methods \cite{wang2020solov2}, \cite{tian2020conditional}. We also tested recently proposed SAM-based approaches, including Efficient-SAM \cite{xiong2024efficientsam} and RSPrompter \cite{chen2024rsprompter}, and methods specifically designed for underwater image instance segmentation, such as USIS-SAM \cite{lian2024diving}. As shown in Table \ref{tab:liaci}, our method achieved the best result in terms of each evaluation metric. In particular, our SPDA-SAM outperformed the second-best approach by 2.6, 1.6 and 5.0 in terms of $mAP$, $AP_{50}$ and $AP_{75}$, respectively, using the mask-level ground-truth data. Given that the box-level ground-truth data was utilized, the three values were 1.3, 2.5 and 2.0, respectively.

\begin{table*}
\centering
\caption{Comparison between our method and ten baseline methods on the LIACI \cite{waszak2022semantic} data set.}
\label{tab:liaci}
\begin{tabular}{c|c|c|ccc|ccc} 
\toprule
\multirow{2}{*}{Method}  & \multirow{2}{*}{Venue} & \multicolumn{1}{l|}{\multirow{2}{*}{Backbone}} & \multicolumn{3}{c|}{Mask}                     & \multicolumn{3}{c}{Box}                        \\
\cline{4-9}
                         &                         & \multicolumn{1}{l|}{}                          & $mAP$         & $AP_{50}$       & $AP_{75}$       & $mAP$         & $AP_{50}$       & $AP_{75}$        \\ 
\hline\hline
Mask R-CNN\cite{he2017mask}               & ICCV17                  & ResNet-50                                      & 36.4          & 58.9          & 40.0          & 36.1          & 60.5          & 38.4           \\
SOLOv2\cite{wang2020solov2}                   & TPAMI21                 & ResNet-50                                      & 34.8          & 54.7          & 37.5          & -             & -             & -              \\
CondInst\cite{tian2020conditional}                 & ECCV20                  & ResNet-50                                      & 38.6          & \underline{60.0}  & 39.7          & \underline{41.2}  & 60.9          & \underline{46.0}   \\
BoxInst\cite{tian2021boxinst}                  & CVPR21                  & ResNet-50                                      & 27.7          & 50.0          & 26.5          & 38.7          & 57.1          & 43.2           \\
FastInst\cite{he2023fastinst}                 & CVPR23                  & ResNet-50                                      & 38.6          & 55.0          & 41.0          & -             & -             & -              \\
HTC\cite{chen2019hybrid}                      & CVPR19                  & ResNet-50                                      & 36.4          & 55.2          & 39.2          & 37.8          & 55.4          & 40.5           \\
QueryInst\cite{fang2021instances}                & CVPR21                  & ResNet-50                                      & 36.0          & 51.0          & 38.9          & 37.0          & 51.0            & 41.0           \\ 
\hline
USIS-SAM\cite{lian2024diving}                 & ICML24                  & ViT-Huge                                       & 27.8          & 44.6          & 28.8          & 26.3          & 47.4            & 25.3           \\
Efficient-SAM\cite{xiong2024efficientsam}          & CVPR24                  & ViT-Small                                      & 33.3          & 51.4          & 33.9          & 35.3          & 54.7            & 39.2           \\
RSPrompter\cite{chen2024rsprompter}               & TGRS24                  & ViT-Base                                       & \underline{40.5}  & 58.7          & \underline{42.3}  & 40.8          & \underline{61.4}  & 45.9           \\
\textbf{SPDA-SAM (Ours)} & -                       & ViT-Base                                       & \textbf{43.1} & \textbf{61.6} & \textbf{47.3} & \textbf{42.5} & \textbf{63.9} & \textbf{48.0}  \\
\bottomrule
\end{tabular}
\end{table*}

\subsubsection{USIS10K}

The results obtained using the proposed SPDA-SAM and ten baselines on the USIS10K \cite{lian2024diving} data set are reported in Table \ref{tab:usis10k}. Following the original benchmark settings, we conducted both multi-class and class-agnostic instance segmentation tasks. Again, our method outperformed its counterparts with regard to each metric. Specifically, our SPDA-SAM achieved the improvements of 4.9, 3.3 and 6.4 in terms of $mAP$, $AP_{50}$ and $AP_{75}$, respectively, compared to the best baseline in the class agnostic instance segmentation task. On the other hand, the three digits were 2.8, 1.3 and 3.8, respectively, within the multi-class instance segmentation task. Compared to the methods which utilized the ViT-Huge backbone \cite{dosovitskiy2020image}, our approach used the more lightweight ViT-Base backbone \cite{dosovitskiy2020image} while still achieving the superior performance.

\begin{table*}
\centering
\caption{Comparison between our method and ten state-of-the-art methods on the USIS10K \cite{lian2024diving} data set.}
\label{tab:usis10k}
\begin{tabular}{c|c|c|ccc|ccc} 
\toprule
\multirow{2}{*}{Method}  & \multirow{2}{*}{Venue} & \multirow{2}{*}{Backbone} & \multicolumn{3}{c|}{Class-Agnostic}           & \multicolumn{3}{c}{Multi-class}                \\
\cline{4-9}
                         &                         &                           & $mAP$         & $AP_{50}$       & $AP_{75}$       & $mAP$         & $AP_{50}$       & $AP_{75}$        \\ 
\hline\hline
WaterMask\cite{lian2023watermask}                & ICCV23                  & ResNet-50                 & 58.3          & 80.2          & 66.5          & 37.7          & 54.0          & 42.5           \\
S4Net\cite{fan2019s4net}                    & CVPR19                  & ResNet-50                 & 32.8          & 64.1          & 27.3          & 23.9          & 43.5          & 24.4           \\
RDPNet\cite{wu2021regularized}                   & TIP21                   & ResNet-50                 & 53.8          & 77.8          & 61.9          & 37.9          & 55.3          & 42.7           \\
OQTR\cite{pei2022transformer}                     & TMM22                   & ResNet-50                 & 56.6          & 79.3          & 62.6          & 19.7          & 30.6          & 21.9           \\ 
\hline
RDPNet\cite{wu2021regularized}                   & TIP21                   & ResNet-101                & 54.7          & 78.3          & 63.0          & 39.3          & 55.9          & 45.4           \\
WaterMask\cite{lian2023watermask}                & ICCV23                  & ResNet-101                & 59.0          & 80.6          & 67.2          & 38.7          & 54.9          & 43.2           \\ 
\hline
RSPrompter\cite{chen2024rsprompter}               & TGRS24                  & ViT-Huge                  & 58.2          & 79.9          & 65.9          & 40.2          & 55.3          & 44.8           \\
SAM+BBox\cite{Kirillov_2023_ICCV}                 & ICCV23                  & ViT-Huge                  & 45.9          & 65.9          & 52.1          & 26.4          & 38.9          & 29.0           \\
SAM+Mask\cite{Kirillov_2023_ICCV}                 & ICCV23                  & ViT-Huge                  & 55.1          & 80.2          & 62.8          & 38.5          & 56.3          & 44.0           \\
USIS-SAM\cite{lian2024diving}                 & ICML24                  & ViT-Huge                  & \underline{59.7}  & \underline{81.6}  & \underline{67.7}  & \underline{43.1}  & \underline{59.0}  & \underline{48.5}   \\
\textbf{SPDA-SAM (Ours)} & -                       & ViT-Base                  & \textbf{64.6} & \textbf{84.9} & \textbf{74.1} & \textbf{45.9} & \textbf{60.3} & \textbf{52.3}  \\
\bottomrule
\end{tabular}
\end{table*}

\subsubsection{UIIS, UIIS10K and USIS16K}
In addition, we evaluated the SPDA-SAM on the UIIS \cite{lian2023watermask}, UIIS10K \cite{li2025uwsam} and USIS16K \cite{hong2025usis16k} data sets. Our method was also compared with six baselines. It can be seen in Table \ref{tab:threedataset}, our SPDA-SAM outperformed the baseline by 3.5, 3.6 and 4.7 in $mAP$, $AP_{50}$ and $AP_{75}$, respectively, on the UIIS data set. When the UIIS10K data set was used, the three values were 4.3, 1.7 and 3.9, respectively. In contrast, the three values were 1.3, 1.7 and 1.2 on the USIS16K data set, respectively. The above results demonstrate the better UIIS capability and the stronger robustness of our SPDA-SAM across diverse underwater scenarios, compared to its counterparts.

\begin{table*}
\centering
\caption{Comparison between our method and six baseline methods on the UIIS \cite{lian2023watermask}, UIIS10K \cite{li2025uwsam} and USIS16K \cite{hong2025usis16k} data sets.}
\label{tab:threedataset}
\begin{tabular}{c|c|ccc|ccc|ccc} 
\toprule
\multirow{2}{*}{Method} & \multirow{2}{*}{Venue} & \multicolumn{3}{c|}{UIIS}                     & \multicolumn{3}{c|}{UIIS10K}                  & \multicolumn{3}{c}{USIS16K}                    \\
\cline{3-11}
                        &                         & $mAP$         & $AP_{50}$       & $AP_{75}$       & $mAP$         & $AP_{50}$       & $AP_{75}$       & $mAP$         & $AP_{50}$       & $AP_{75}$        \\ 
\hline\hline
Mask R-CNN\cite{he2017mask}              & ICCV17                   & 23.4          & 40.9          & 25.3          & 35.8          & 53.6          & 40.2          & 73.6          & 90.0          & 81.7           \\
Mask Scoring R-CNN\cite{huang2019mask}      & CVPR19                  & 24.6          & 41.9          & 26.5          & 37.3          & \underline{53.7}          & \underline{41.8}          & 71.6          & 88.5          & 79.8           \\
PointRend\cite{kirillov2020pointrend}               & CVPR20                  & 25.9          & 43.4          & 27.6          & 37.3          & 53.0          & 41.3          & 76.3          & 89.4          & 82.7           \\
WaterMask\cite{lian2023watermask}               & ICCV23                  & \underline{27.2}          & \underline{43.7}          & \underline{29.3}          & \underline{37.4}          & 51.6          & 41.7          & 72.7          & 86.8          & 79.3           \\
RSPrompter\cite{chen2024rsprompter}              & TGRS24                  & 25.1          & 40.3          & 26.2          & 33.2          & 44.9          & 36.0          & 74.6          & 86.4          & 82.1           \\
USIS-SAM\cite{lian2024diving}                & ICML24                  & 26.3          & 41.3          & 28.7          & 35.8          & 53.6          & 40.6          & \underline{81.0}          & \underline{90.8}          & \underline{87.1}           \\
\textbf{SPDA-SAM (Ours)}         & -                       & \textbf{30.7} & \textbf{47.3} & \textbf{34.0} & \textbf{41.7} & \textbf{55.4} & \textbf{45.7} & \textbf{82.3} & \textbf{92.5} & \textbf{88.3}  \\
\bottomrule
\end{tabular}
\end{table*}

\subsubsection{ZeroWaste}

The ZeroWaste \cite{bashkirova2022zerowaste} data set was collected in extremely cluttered recycling environments. We used it to evaluate the robustness of our SPDA-SAM in handling complex object interactions and diverse material appearances.
The results obtained using our SPDA-SAM and nine baselines on the ZeroWaste data set are shown in Table \ref{tab:zerowaste}. It can be observed that SPDA-SAM achieved the best performance in terms of the $mAP$ and $AP_{75}$ metrics, which reach 34.4 and 38.2, respectively. Although USIS-SAM (ViT-Huge) \cite{lian2024diving} obtained the highest $AP_{50}$ value, SPDA-SAM utilized a lighter ViT-Base backbone and outperformed the other baselines with regard to this metric. These results indicate that the proposed SPDA-SAM is able to perform robust and competitive instance segmentation in highly cluttered real-world scenarios.

\begin{table*}
\centering
\caption{Comparison between our SPDA-SAM and nine state-of-the-art methods on the ZeroWaste \cite{bashkirova2022zerowaste} data set. Except the three overall evaluation metrics, per-class $mAP$ was also computed for each object category.}
\label{tab:zerowaste}
\begin{tabular}{c|c|c|ccc|cccc} 
\toprule
Method                                              & Venue   & Backbone  & $mAP$         & $AP_{50}$     & $AP_{75}$     & Rigid\_Plastic & Cardboard     & Metal         & Soft\_Plastic  \\ 
\hline\hline
Mask R-CNN \cite{he2017mask}                             & ICCV17  & ResNet-50 & 29.0          & 44.9          & 31.8          & 25.5           & 32.8          & 24.4          & 33.2           \\
SOLOv2 \cite{wang2020solov2}                                & TPAMI21 & ResNet-50 & 19.4          & 35.9          & 17.7          & 11.7           & 27.5          & 15.3          & 22.9           \\
CondInst \cite{tian2020conditional}                          & ECCV20  & ResNet-50 & 30.4          & 45.2          & 33.2          & \underline{30.0}   & 35.9          & 23.9          & 31.8           \\
BoxInst \cite{tian2021boxinst}                                            & CVPR21  & ResNet-50 & 26.5          & 42.0          & 29.2          & 21.5           & 33.1          & 23.9          & 27.6           \\
FastInst \cite{he2023fastinst}                                            & CVPR23  & ResNet-50 & 27.5          & 40.4          & 29.4          & 24.4           & 29.7          & 24.9          & 30.9           \\
HTC \cite{chen2019hybrid}                                                 & CVPR19  & ResNet-50 & 29.2          & 46.8          & 31.9          & 24.3           & 34.1          & \underline{25.4}  & 32.9           \\
QueryInst \cite{fang2021instances}                                           & CVPR21  & ResNet-50 & 26.1          & 40.2          & 28.6          & 24.3           & 29.5          & 21.1          & 29.4           \\ 
\hline
RSPrompter \cite{chen2024rsprompter}                                         & TGRS24  & ViT-Base  & 28.7          & 41.7          & 32.0          & 22.5           & 39.3          & 16.6          & 36.2           \\
USIS-SAM \cite{lian2024diving}                                           & ICML24  & ViT-Huge  & \underline{33.8}  & \textbf{49.8} & \underline{37.6}  & 26.8           & \textbf{43.2} & \textbf{26.1} & \underline{39.1}   \\
\textbf{\textbf{\textbf{\textbf{SPDA-SAM (Ours)}}}} & -       & ViT-Base  & \textbf{34.4} & \underline{47.8}  & \textbf{38.2} & \textbf{34.5}  & \underline{41.5}  & 21.5          & \textbf{39.8}  \\
\bottomrule
\end{tabular}
\end{table*}

\subsubsection{Computational Complexity Analysis}

We further compared the proposed SPDA-SAM with ten instance segmentation methods in computational complexity, as shown in Table \ref{tab:complexity}. It can be seen that our SPDA-SAM maintained a proper inference speed with a moderate number of parameters and FLOPs using the ViT-Base \cite{dosovitskiy2020image} backbone. Although the SPDA-SAM incurred the higher FLOPs value than lightweight models, such as FastInst \cite{he2023fastinst} and Efficient-SAM \cite{xiong2024efficientsam}, it achieved the better segmentation performance. As shown in Table \ref{tab:liaci}, our method outperformed FastInst and Efficient-SAM by 4.5 and 9.8, respectively, in terms of the $mAP$ metric under the mask-level ground-truth setting. Compared to USIS-SAM \cite{lian2024diving}, our SPDA-SAM reduced parameters and FLOPs by more than half while improving the mask $mAP$ value from 27.8 to 43.1 and the box $mAP$ value from 26.3 to 42.5. It is suggested that our SPDA-SAM achieved a proper balance between computational complexity and segmentation accuracy.

\begin{table}
\centering
\setlength{\tabcolsep}{2pt}
\caption{Comparison between our SPDA-SAM and nine instance segmentation methods in terms of the number of parameters (M), FLOPs (G) and inference speed (FPS).}
\label{tab:complexity}
\begin{tabular}{c|c|c|c|c} 
\toprule
Method                   & Venue   & Params (M) & FLOPs (G) & Speed (FPS)  \\ 
\hline\hline
Mask R-CNN\cite{he2017mask}        & ICCV17  & 44.4       & 253.0     & 13.8         \\
SOLOv2\cite{wang2020solov2}        & TPAMI21 & 46.6       & 239.0     & 13.4         \\
CondInst\cite{tian2020conditional}         & ECCV20  & 34.2       & 331.0     & 22.7         \\
BoxInst\cite{tian2021boxinst}       & CVPR21  & 35.1       & 372.0     & 22.0         \\
FastInst\cite{he2023fastinst}       & CVPR23  & 34.1       & 58.2      & 28.9         \\
QueryInst\cite{fang2021instances}       & CVPR21  & 172.0      & 170.0     & 14.5         \\
PolySnake\cite{feng2024recurrent}      & TCSVT24 & 22.0       & 308.6     & 3.5          \\ 
\hline
RSPrompter\cite{chen2024rsprompter}     & TGRS24  & 117.5      & 114.5     & 15.5         \\
USIS-SAM\cite{lian2024diving}     & ICML24  & 696.8      & 824.4     & 5.6          \\
\textbf{SPDA-SAM (Ours)} & -       & 225.37     & 264.5     & 12.2         \\
\bottomrule
\end{tabular}
\end{table}

\subsection{Ablation Study}

We performed three ablation experiments to examine the impacts of different modules of our SPDA-SAM. 
For simplicity, only the LIACI \cite{waszak2022semantic}, ZeroWaste\cite{bashkirova2022zerowaste} and USIS10K \cite{lian2024diving} data set were used. Regarding the LIACI data set, its mask-level ground-truth was utilized.

\subsubsection{Impact of the C2FFM}

To investigate the effectiveness of the proposed C2FFM, we removed it from the SPDA-SAM. In this case, only the RGB images were used. As reported in Table \ref{tab:ablation_C2FFM}, the incorporation of the C2FFM into our SPDA-SAM led to an increase in performance across the three data sets regardless which evaluation metric was used. It is also indicated that the depth information is useful for the instance segmentation task, in particular, when used with our method.


\begin{table*}
\centering
\caption{Comparison between the results obtained using the SPDA-SAM with and without the C2FFM.}
\label{tab:ablation_C2FFM}
\begin{tabular}{c|ccc|ccc|ccc} 
\toprule
\multirow{2}{*}{Method} & \multicolumn{3}{c|}{LIACI}                    & \multicolumn{3}{c|}{ZeroWaste}                & \multicolumn{3}{c}{USIS10K}                    \\
\cline{2-10}
                        & $mAP$         & $AP_{50}$     & $AP_{75}$     & $mAP$         & $AP_{50}$     & $AP_{75}$     & $mAP$         & $AP_{50}$     & $AP_{75}$      \\ 
\hline\hline
w/o C2FFM               & 42.0          & 59.7          & 46.3          & 33.2          & 45.9          & 37.2          & 44.6          & 59.5          & 51.5           \\
w/ C2FFM                & \textbf{43.1} & \textbf{61.5} & \textbf{47.3} & \textbf{34.4} & \textbf{47.8} & \textbf{38.2} & \textbf{45.9} & \textbf{60.3} & \textbf{52.3}  \\
\bottomrule
\end{tabular}
\end{table*}

\subsubsection{Impact of the SSSPM}


We further examined the effect of the proposed SSSPM on our SPDA-SAM. Six variants of the SPDA-SAM were obtained by removing the entire SSSPM, using only the spatial prompt or the semantic prompt, and adopting different fusion strategies for the semantic and spatial prompts.  These variants were compared with the SPDA-SAM in Table \ref{tab:ablation_SSSPM}. It can be observed that the use of the entire SSSPM produced the best result across the three data sets no matter which metric was used. It is suggested that the self-prompt module is useful for our SPDA-SAM.

\begin{table*}
\centering
\renewcommand\arraystretch{1.2}
\setlength{\tabcolsep}{4pt}
\caption{Comparison of the results obtained using the SPDA-SAM with different prompt methods.}
\label{tab:ablation_SSSPM}
\begin{tabular}{c|c|ccc|ccc|ccc} 
\toprule
\multirow{2}{*}{Prompt Method}    & \multirow{2}{*}{Fusion Method} & \multicolumn{3}{c|}{LIACI}                    & \multicolumn{3}{c|}{ZeroWaste}                & \multicolumn{3}{c}{USIS10K}                    \\
\cline{3-11}
                                  &                         & $mAP$         & $AP_{50}$     & $AP_{75}$     & $mAP$         & $AP_{50}$     & $AP_{75}$     & $mAP$         & $AP_{50}$     & $AP_{75}$      \\ 
\hline\hline
w/o SSSPM                         & -                       & 41.0          & 58.8          & 44.4          & 32.6          & 45.4          & 35.7          & 44.5          & 58.8          & 50.3           \\
w/ Semantic                       & -                       & 41.6          & 59.2          & \underline{46.0}  & 33.3          & 45.8          & 37.0          & \underline{45.0}  & \underline{60.0}  & 51.3           \\
w/ Spatial                        & -                       & 42.1          & 60.3          & 45.8          & 33.1          & 45.9          & 36.4          & \underline{45.0}  & 59.8          & 50.5           \\ 
\hline
\multirow{3}{*}{w/ Semantic \& Spatial} & Addition                     & 40.8          & 58.0          & 43.9          & 32.2          & 45.0          & 35.3          & 43.7          & 57.0          & 48.7           \\
                                  & Cross-Attention         & \underline{42.3}  & \underline{61.0}  & 45.9          & \underline{33.5}  & \underline{46.2}  & \underline{37.2}  & 44.8          & 59.8          & \underline{51.5}   \\
                                  & Prompt Fusion Block     & \textbf{43.1} & \textbf{61.5} & \textbf{47.3} & \textbf{34.4} & \textbf{47.8} & \textbf{38.2} & \textbf{45.9} & \textbf{60.3} & \textbf{52.3}  \\
\bottomrule
\end{tabular}
\end{table*}

\subsubsection{Impact of the Depth Estimator}

\begin{figure}
  \centering
  \includegraphics[width=1.0\linewidth]{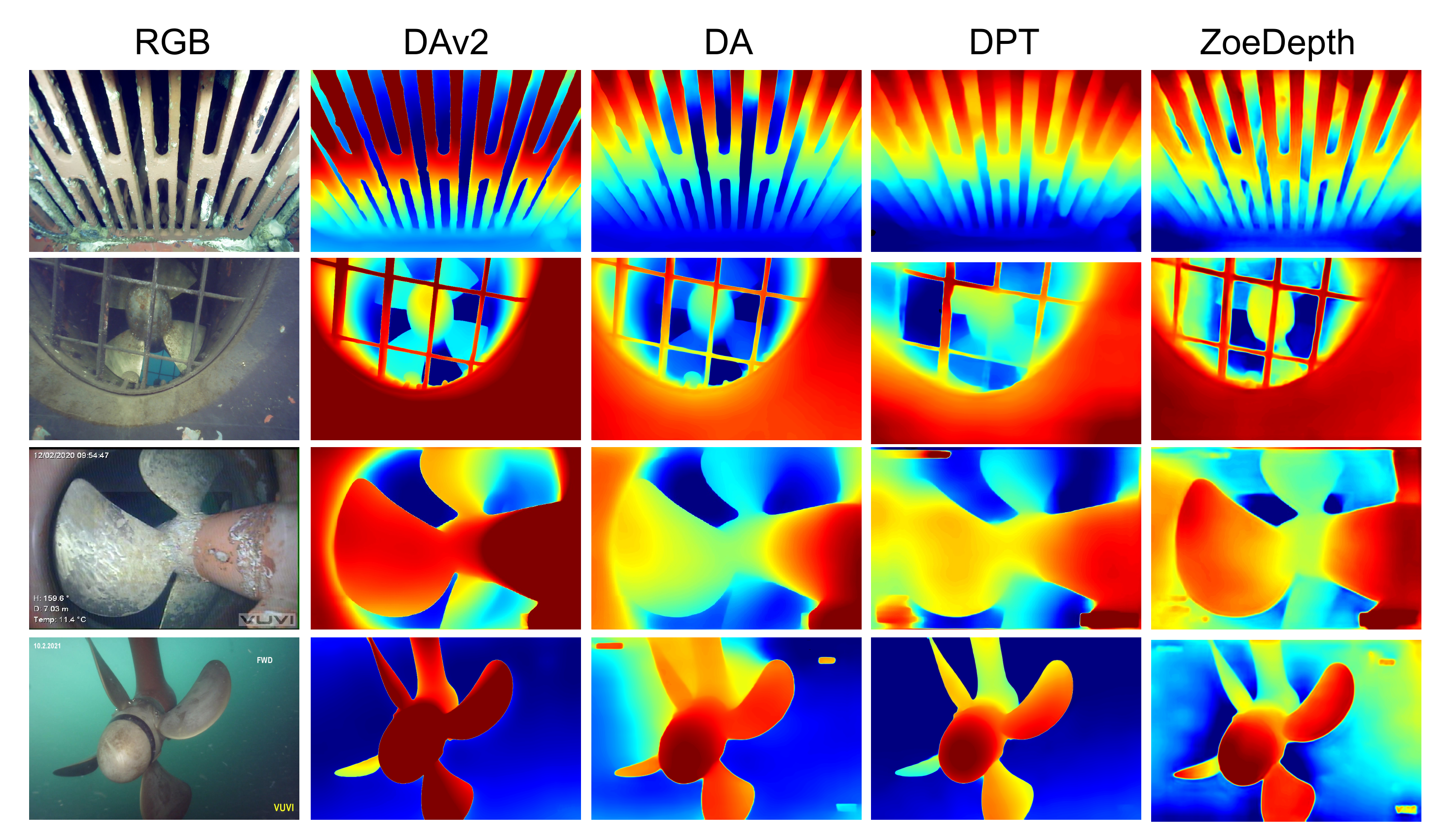}
  \caption{Visual comparison of the depth maps obtained using four depth estimators from four underwater hull images in the LIACI \cite{waszak2022semantic} data set.
  }
  \label{fig:ablation_depth_vis}
\end{figure}

In addition, we investigated the effect of the depth estimator on our method. Four state-of-the-art monocular depth estimators were tested, including Dense Prediction Transformer (DPT) \cite{ranftl2021vision}, Zero-shot Transfer by Combining Relative and Metric Depth (ZoeDepth) \cite{bhat2023zoedepth}, Depth Anything (DA) \cite{depthanything} and  Depth Anything v2 (DAv2) \cite{yang2024depth}. Given four underwater hull images, the depth maps that the four depth estimators generated were visualized in Fig. \ref{fig:ablation_depth_vis}. It can be seen that the depth maps produced by DAv2 manifest the clearer boundaries, more consistent structures and better depth layering, compared with the other depth estimators. The results derived using our SPDA-SAM together with the four depth estimators are also compared in Table \ref{tab:ablation_depth_liaci}. As can be seen, the utilization of DAv2 normally produced the better result than those derived using the other depth estimators across the three data sets, regardless of which metric was used. This finding demonstrates that the quality of the depth map obtained using a pre-trained estimator is important to the performance of our SPDA-SAM.


Notably, the results in Table~\ref{tab:ablation_C2FFM}, together with those in Tables~\ref{tab:liaci} and~\ref{tab:usis10k}, show that our method consistently outperformed the baselines across the three datasets even when depth information was unavailable.

\begin{table*}
\centering
\caption{Comparison of the results obtained using the SPDA-SAM with different depth estimators.}
\label{tab:ablation_depth_liaci}
\begin{tabular}{c|ccc|ccc|ccc} 
\toprule
\multirow{2}{*}{Depth Estimator} & \multicolumn{3}{c|}{LIACI}                    & \multicolumn{3}{c|}{ZeroWaste}                & \multicolumn{3}{c}{USIS0K}                     \\
\cline{2-10}
                                 & $mAP$         & $AP_{50}$     & $AP_{75}$     & $mAP$         & $AP_{50}$     & $AP_{75}$     & $mAP$         & $AP_{50}$     & $AP_{75}$      \\ 
\hline\hline
Dense Prediction Transformer     & 42.4          & 58.3          & 47.0          & 32.5          & 45.1          & 35.9          & 44.8          & 59.5          & 51.6           \\
ZoeDepth                         & 42.2          & 59.1          & 46.8          & 32.9          & 45.6          & 36.7          & 44.6          & 59.4          & 51.1           \\
Depth Anything                   & \underline{42.5}  & \underline{60.5}  & \underline{47.2}  & \underline{33.6}  & \underline{46.6}  & \underline{37.0}  & \underline{45.3}  & \underline{59.8}  & \underline{51.9}   \\
Depth Anything v2                & \textbf{43.1} & \textbf{61.5} & \textbf{47.3} & \textbf{34.4} & \textbf{47.8} & \textbf{38.2} & \textbf{45.9} & \textbf{60.3} & \textbf{52.3}  \\
\bottomrule
\end{tabular}
\end{table*}

\subsubsection{Impact of the Backbone}

To assess the effect of different backbones, we replaced the SAM \cite{Kirillov_2023_ICCV} backbone in SPDA-SAM with that of SAM2 \cite{ravi2025sam} and SAM3 \cite{carion2025sam} separately while keeping the rest of SPDA-SAM unchanged. The results are reported in Table~\ref{tab:ablation_backbone}. As can be observed, the SPDA-SAM with the SAM backbone achieved the best performance on the three data sets. This finding suggests that the proposed depth-aware dual-path encoder and self-prompted decoder are better aligned with the feature representation of the image encoder of SAM. In contrast, the backbones of SAM2 \cite{ravi2025sam} and SAM3 \cite{carion2025sam} may introduce different feature distributions and architectural priors, which require additional adaptation to fully exploit their representation ability. 

\begin{table*}
\centering
\caption{Comparison between SPDA-SAM and its two variants obtained by replacing the SAM backbone with that of later SAMs.}
\label{tab:ablation_backbone}
\begin{tabular}{c|ccc|ccc|ccc} 
\toprule
\multirow{2}{*}{Backbone} & \multicolumn{3}{c|}{LIACI}                    & \multicolumn{3}{c|}{ZeroWaste}                & \multicolumn{3}{c}{USIS10K}                    \\
\cline{2-10}
                                 & $mAP$         & $AP_{50}$     & $AP_{75}$     & $mAP$         & $AP_{50}$     & $AP_{75}$     & $mAP$         & $AP_{50}$     & $AP_{75}$      \\ 
\hline\hline
SAM3 \cite{carion2025sam}                             & \underline{39.5}          & 60.6          & \underline{44.1}          & \underline{31.5}          & \underline{45.5}          & \underline{34.2}          & \underline{43.8}          & \underline{59.5}          & \underline{49.3}           \\
SAM2 \cite{ravi2025sam}                             & 39.0          & \underline{61.0}          & 43.1          & 30.2          & 45.1          & 32.8          & 43.2          & 59.3          & 48.8           \\
SAM\cite{Kirillov_2023_ICCV}                             & \textbf{43.1} & \textbf{61.5} & \textbf{47.3} & \textbf{34.4} & \textbf{47.8} & \textbf{38.2} & \textbf{45.9} & \textbf{60.3} & \textbf{52.3}  \\
\bottomrule
\end{tabular}
\end{table*}

\subsection{Visualization}

In this subsection, we will visualize the results obtained using the proposed SPDA-SAM, to intuitively convey the information that our method produced.

\subsubsection{Visualization of Instance Segmentation Masks}

\begin{figure*}
  \centering
  \includegraphics[width=1.0\linewidth]{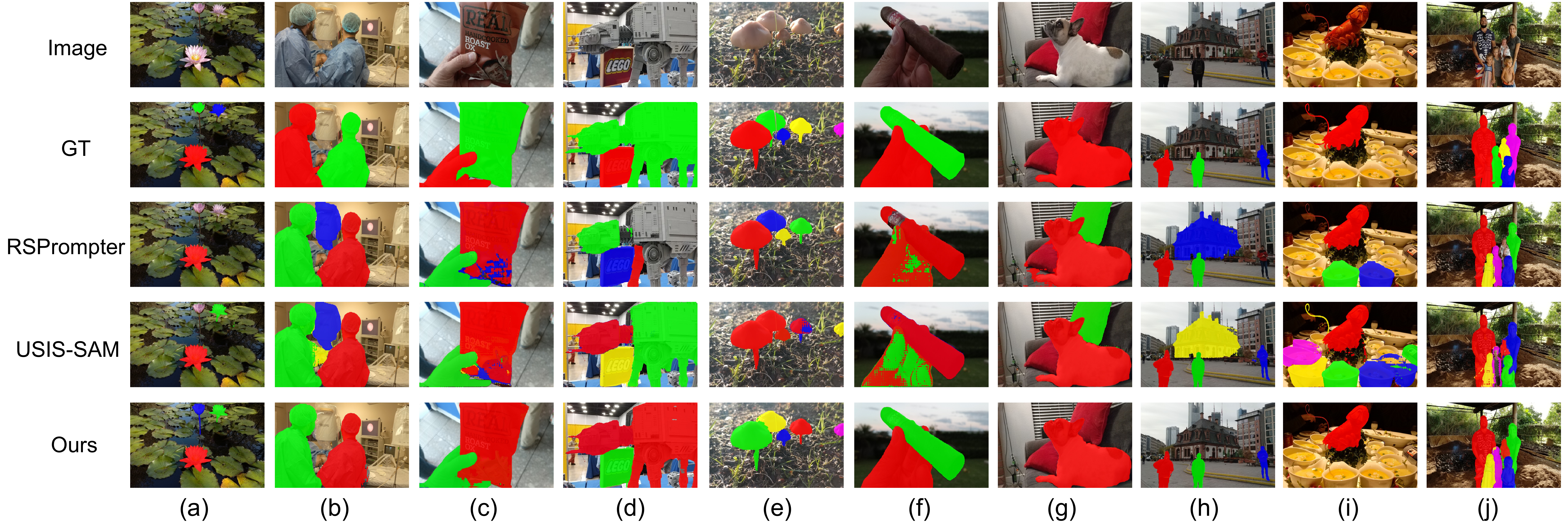}
  \caption{Visualization of the results derived using two state-of-the-art baselines, i.e., RSPrompter \cite{chen2024rsprompter} and USIS-SAM \cite{lian2024diving}, and our method on the 10 images in the COME15K-E \cite{zhang2021rgb} data set.}
  \label{fig:visualization_COME-E}
\end{figure*}

\begin{figure*}
  \centering
  \includegraphics[width=1.0\linewidth]{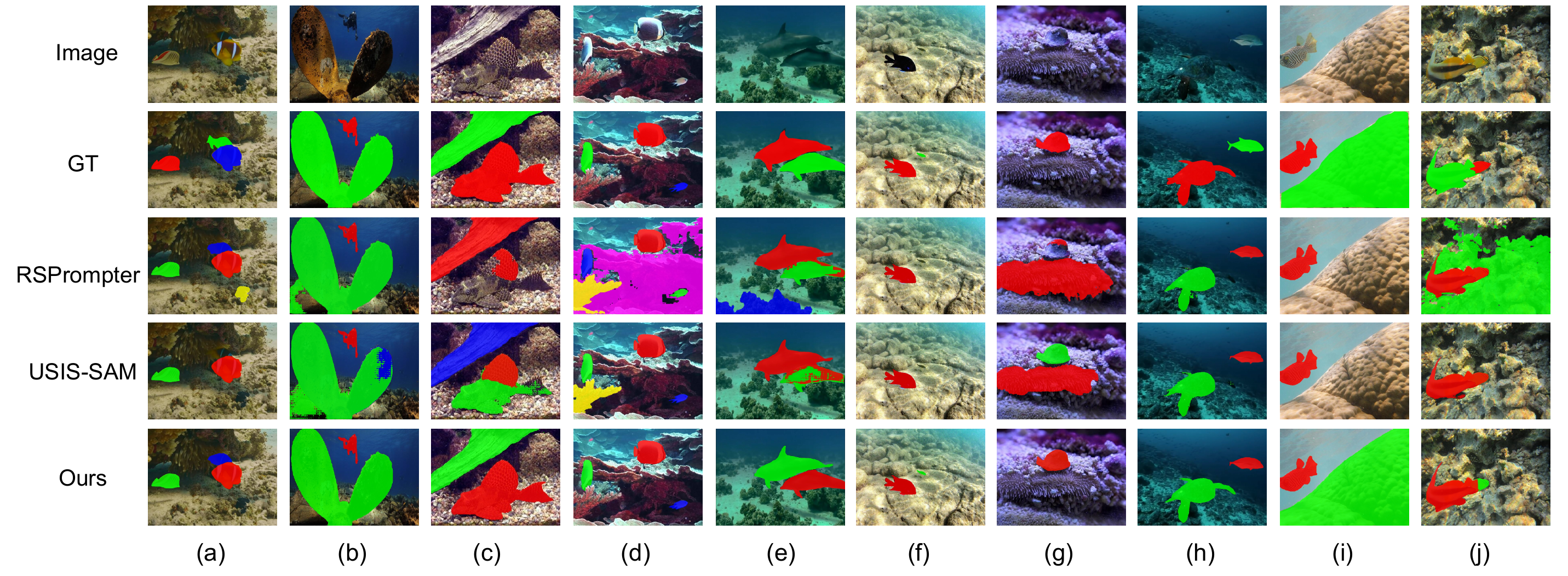}
  \caption{Visualization of the results derived using two state-of-the-art baselines, i.e., RSPrompter \cite{chen2024rsprompter} and USIS-SAM \cite{lian2024diving}, and our method on the 10 images in the USIS10K \cite{lian2024diving} data set.}
  \label{fig:visualization_USIS10K}
\end{figure*}

We visualize the results produced by two state-of-the-art approaches, including RSPrompter \cite{chen2024rsprompter} and USIS-SAM \cite{lian2024diving}, and our method on the COME15K-E \cite{zhang2021rgb} and USIS10K \cite{lian2024diving} data sets in Figs. \ref{fig:visualization_COME-E} and \ref{fig:visualization_USIS10K}, respectively. As shown in Fig.~\ref{fig:visualization_COME-E}, our SPDA-SAM generated more accurate and complete instance masks (see (c), (d) and (f)), detected a more complete set of instance objects (see (a), (e) and (j)), and better distinguished foreground objects from the background ((b), (g), (h) and (i)), compared to RSPrompter and USIS-SAM. Similarly, our method achieved the performance superior to its counterparts in handling complex scenes and object boundaries on the USIS10K \cite{lian2024diving} data set, as presented in Fig.~\ref{fig:visualization_USIS10K}. These findings should be due to the fact that our method utilized depth information, except the RGB image characteristics that its counterparts also used. 


\subsubsection{Visualization of Intermediate Results}

\begin{figure*}[t]
  \centering
  \includegraphics[width=1.0\linewidth]{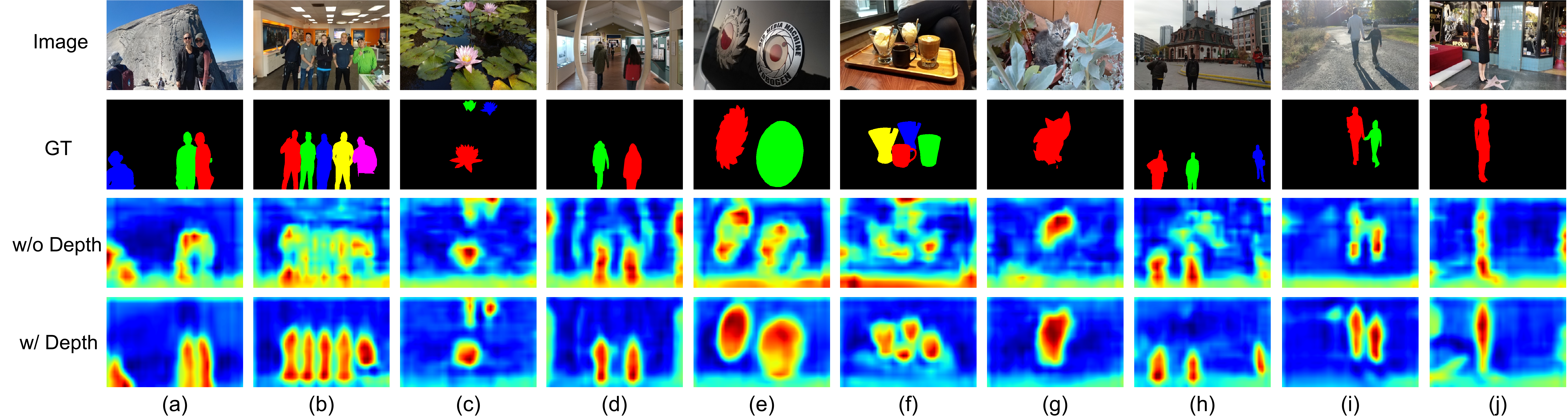}
  \caption{Visualization of the complementary actions between the depth information and original image characteristics. With regard to an RGB image, the image embeddings generated by the depth-aware dual-path encoder are shown.
  }
  \label{fig:analysis}
\end{figure*}

To intuitively understand the complementary actions between the depth information and the original image characteristics, we visualize the image embeddings produced by our depth-aware dual-path encoder in Fig.~\ref{fig:analysis}. It can be seen that object instances become more distinguishable and exhibit stronger contrast against the surrounding background once the depth data has been injected into the network, compared to those obtained without using depth information.

\begin{figure}[t]
  \centering
  \includegraphics[width=1.0\linewidth]{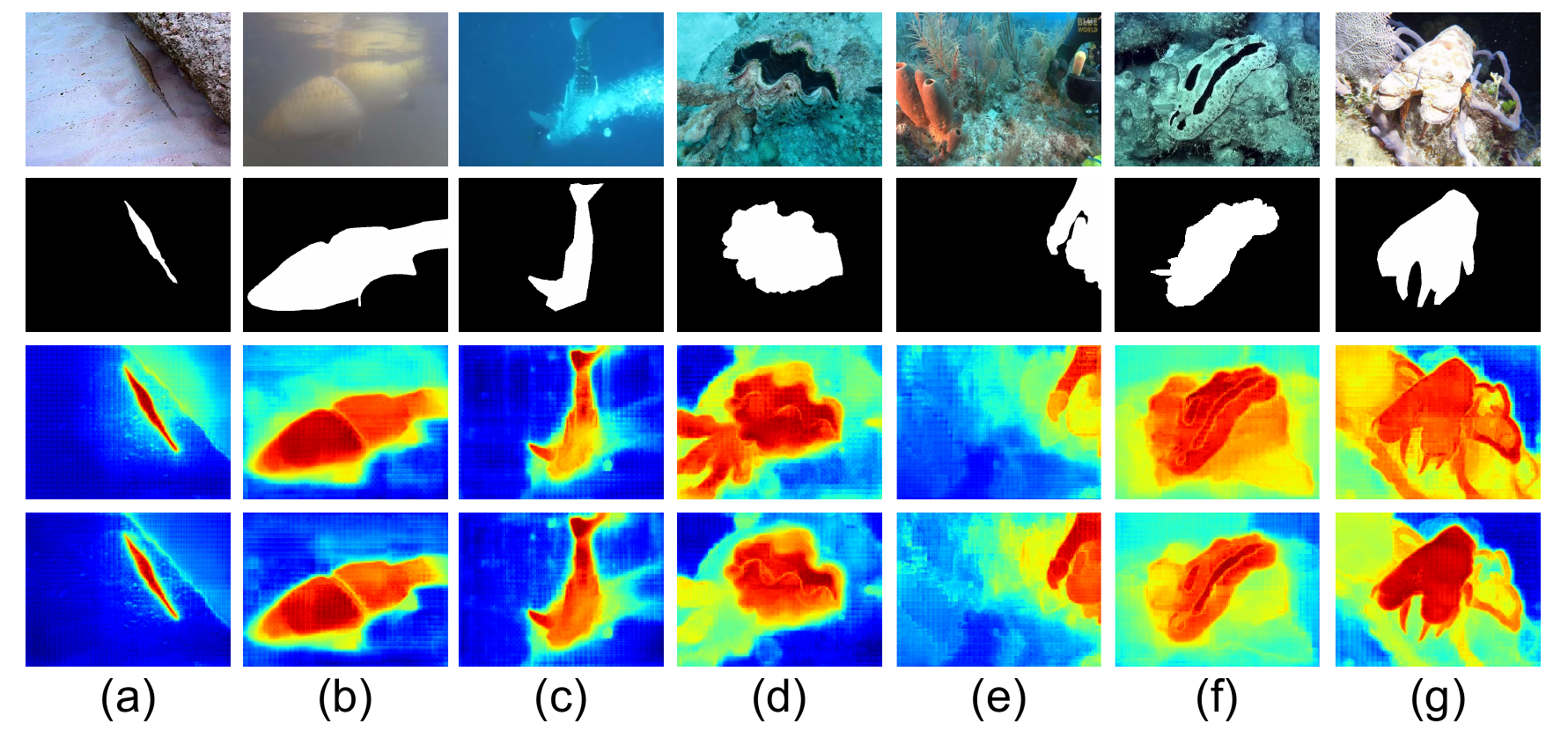}
  \caption{Visualization of the foreground probability maps produced using the mask decoder of the SSSPM during the two-stage segmentation process. Within each group, the input image, ground-truth mask and two foreground probability maps predicted by the mask decoder in its first and second invocations are shown in turn.
  }
  \label{fig:foreground_map}
\end{figure}

In addition, the foreground probability maps produced by the mask decoder of the SSSPM are visulized in Fig. \ref{fig:foreground_map}, to intuitively demonstrate the effectiveness of the self-prompt mechanism. As can be seen, the foreground probability map predicted by the mask decoder in its second invocation with self-prompt shows higher confidences in the foreground regions, compared to that predicted in the first invocation. It is suggested that the self-prompt mechanism used by our SSSPM is useful for improving the performance of SAM.

\section{Conclusion}
In this paper, we proposed a Self-prompted Depth-Aware Segment Anything Model (SPDA-SAM) for instance segmentation. Specifically, we designed a Semantic-Spatial Self-prompt Module (SSSPM) to alleviate the reliance of SAM on manual prompts. This module fused the high-level semantic features extracted from the encoder and the features extracted from the preliminary mask that the mask decoder generated. As a result, a prompt enriched by both semantic and spatial information was produced, which could direct the segmentation process. This design freed SAM from the reliance on human intervention. For the purpose of addressing the absence of depth information in instance segmentation, we introduced a Coarse-to-Fine RGB-D Fusion Module (C2FFM), which progressively performed coarse-to-fine feature fusion across different stages in our dual-path encoder. This module enabled effective exploitation of the complementary information embedded in RGB images and depth maps. To the authors' knowledge, SAM has not been explored in such self-prompted and depth-aware manners. Experimental results showed that our SPDA-SAM normally outperformed its state-of-the-art counterparts across 11 instance segmentation data sets. We believe that the superior results are due to the guidance of the self-prompts and the compensation for the spatial information loss benefited from the coarse-to-fine RGB-D fusion.

\bibliographystyle{IEEEtran}
\bibliography{refs}

 
\vspace{-30pt}


\begin{IEEEbiography}[{\includegraphics[width=1in,height=1.25in,clip,keepaspectratio]{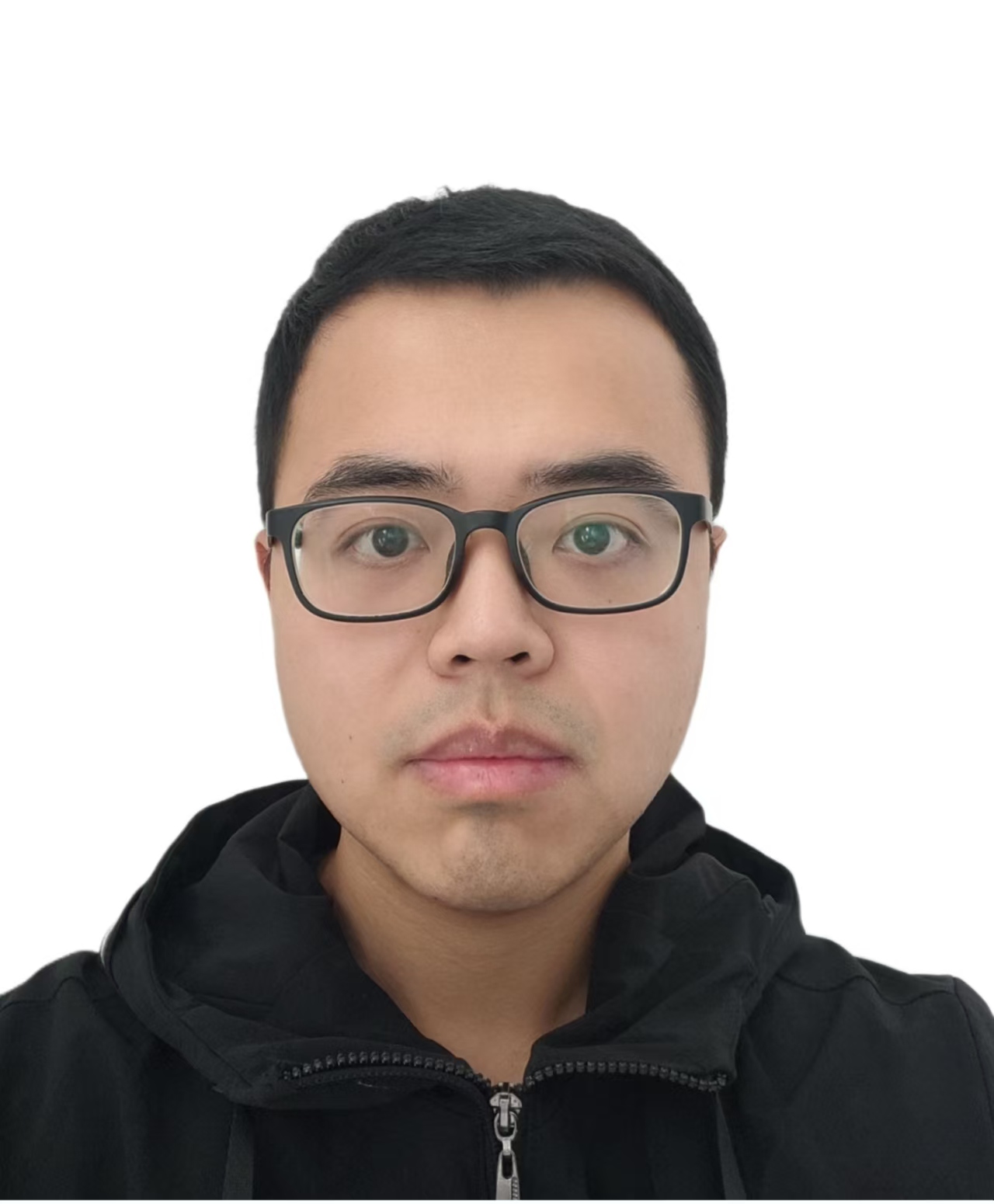}}]{Yihan Shang}
received the B.S. degree in Network Engineering from Harbin University of Science and Technology, Harbin, China, in 2021. He is currently pursuing the M.S. degree in Computer Technology with the Ocean University of China, Qingdao, China. His research interests include computer vision, image segmentation.
\end{IEEEbiography}

\vspace{-30pt}

\begin{IEEEbiography}
[{\includegraphics[width=1in,height=1.25in,clip,keepaspectratio]{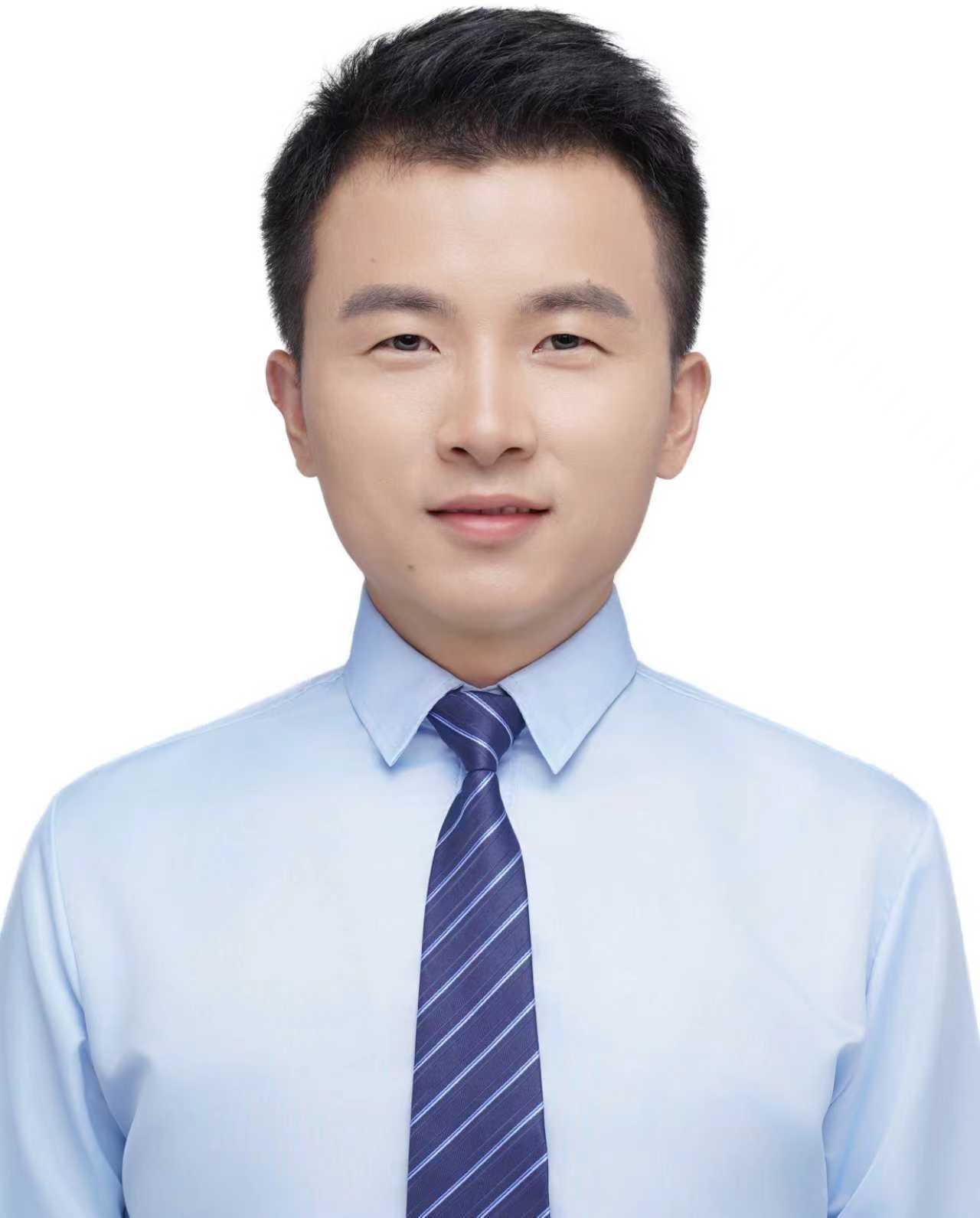}}]{Wei Wang} is currently a post-doctoral candidate at the School of Cyber Science and Technology, Shenzhen Campus of Sun Yat-sen university, Shenzhen, China. He received the Ph.D. degree at the School of Software Technology, Dalian University of Technology, Dalian, China, in 2022. He received the M.S. degree at the School of Computer Science and Technology from the Anhui University, Hefei, China, in 2018. His major research interests include transfer learning, zero-shot learning, deep learning, etc.
\end{IEEEbiography}

\vspace{-30pt}

\begin{IEEEbiography}
[{\includegraphics[width=1in,height=1.25in,clip,keepaspectratio]{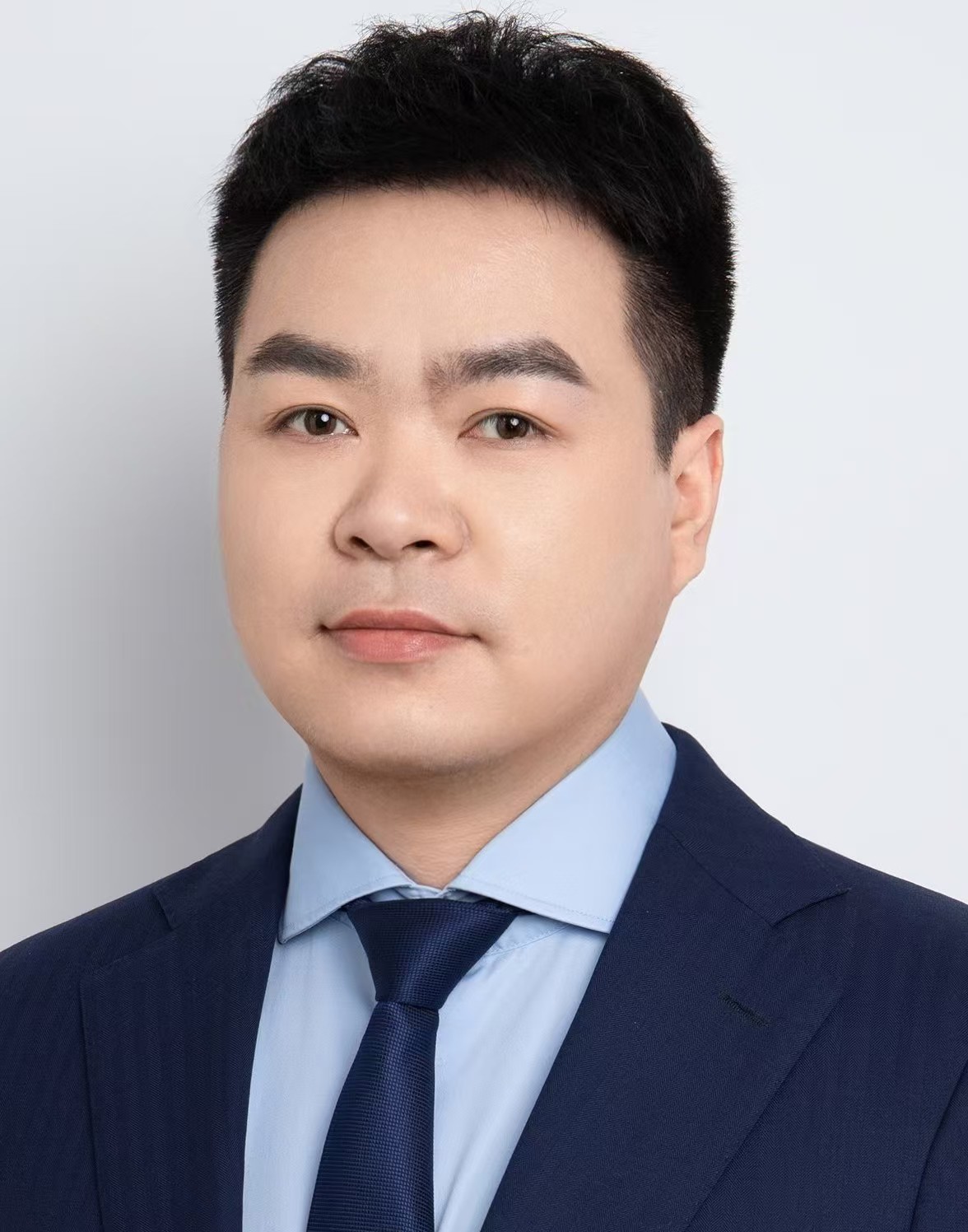}}]{Chao Huang} (Member, IEEE) received the Ph.D. degree in computer science and technology from Harbin Institute of Technology, Shenzhen, China, in 2022. From 2019 to 2022, he was a visiting scholar with Peng Cheng Laboratory, Shenzhen. He is currently an Assistant Professor with the School of Cyber Science and Technology, Sun Yat-sen University, Shenzhen. So far, he has published over 60 technical papers in prestigious international journals and conferences. His research interests include anomaly detection, multimedia analysis, object detection, image/video compression, and deep learning. Dr. Huang received the Distinguished Paper Award of AAAI 2023, and his dissertation was nominated for Harbin Institute of Technology’s Outstanding Dissertation Award. He serves as an Associate Editor for Pattern Recognition and serves/served as the reviewer/ PC member for several top-tier journals and conferences, including IEEE TPAMI, TIP, TIFS, TNNLS, ACM CSUR, CVPR, ICCV, ECCV, ICML, NeurIPs, ICLR, AAAI, IJCAI, and ACM Multimedia.
\end{IEEEbiography}

\vspace{-30pt}

\begin{IEEEbiography}
[{\includegraphics[width=1in,height=1.25in,clip,keepaspectratio]{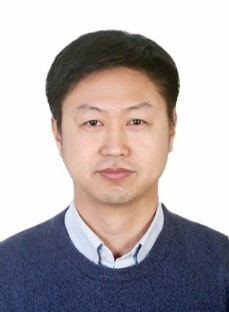}}]
{Junyu Dong} received the B.Sc. and M.Sc. degrees from the Department of Applied Mathematics, Ocean University of China, Qingdao, China, in 1993 and 1999, respectively, and the Ph.D. degree in
image processing from the Department of Computer Science, Heriot-Watt University, U.K., in 2003. He joined Ocean University of China in 2004. He is currently a Professor and the Dean of the Faculty of Information Science and Engineering, Ocean University of China. His research interests include computer vision, underwater image processing, and machine learning, with more than ten research projects supported by the
NSFC, MOST, and other funding agencies.

\end{IEEEbiography}

\vspace{-30pt}

\begin{IEEEbiography}
[{\includegraphics[width=1in,height=1.25in,clip,keepaspectratio]{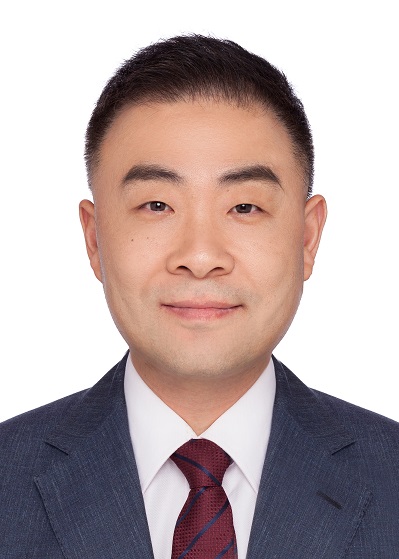}}]{Xinghui Dong} (Member, IEEE) received the Ph.D.degree from Heriot-Watt University, U.K., in 2014. He was with the Centre for Imaging Sciences, University of Manchester, U.K., from 2015 to 2021. In 2021, he joined the Ocean University of China, where he is currently a Professor. His research interests include computer vision, defect detection, texture analysis, and visual perception.
\end{IEEEbiography}



\end{document}